\documentclass[runningheads]{llncs}

\def\paperTitle{Joint-Embedding Predictive Architecture for Self-Supervised Learning of Mask Classification Architecture}

\usepackage{microtype}
\usepackage{epsfig}
\usepackage{caption}
\usepackage{float}
\usepackage{placeins}
\usepackage{color, colortbl}
\usepackage{stfloats}
\usepackage{enumitem}
\usepackage{tabularx}
\usepackage{xstring}
\usepackage{multirow}
\usepackage{xspace}
\usepackage{subcaption}

\usepackage{graphicx}
\usepackage{amsmath}
\usepackage{arydshln}
\usepackage[utf8]{inputenc} 
\usepackage[T1]{fontenc}    
\usepackage{booktabs}       
\usepackage{amsfonts}       
\usepackage{nicefrac}       
\usepackage{microtype}      
\usepackage{adjustbox}
\usepackage{wrapfig}

\usepackage{bbding}
\usepackage{multirow}
\usepackage{booktabs}
\usepackage{makecell}
\usepackage{bbm}
\usepackage{courier}
\usepackage{subfloat}
\usepackage{algorithmic}
\usepackage{tablefootnote}
\usepackage{tabulary,multirow,overpic,subfloat}
\usepackage[hang,flushmargin]{footmisc}

% ...

\newcommand{\app}{\raise.17ex\hbox{$\scriptstyle\sim$}}

\newcolumntype{x}[1]{>{\centering\arraybackslash}p{#1pt}}
\newcolumntype{y}[1]{>{\raggedright\arraybackslash}p{#1pt}}
\newcommand{\dt}[1]{\fontsize{5pt}{0.1em}\selectfont (#1)}
\newlength\savewidth\newcommand\shline{\noalign{\global\savewidth\arrayrulewidth
  \global\arrayrulewidth 1pt}\hline\noalign{\global\arrayrulewidth\savewidth}}
\newcommand{\tablestyle}[2]{\setlength{\tabcolsep}{#1}\renewcommand{\arraystretch}{#2}\centering\footnotesize}

\newcommand{\nbf}[1]{{\noindent \textbf{#1}}}

\newcommand{\R}[1]{{%
    \textbf{%
        \ifstrequal{#1}{1}{\textcolor{red}{R#1}}{%
        \ifstrequal{#1}{2}{\textcolor{blue}{R#1}}{%
        \ifstrequal{#1}{3}{\textcolor{magenta}{R#1}}{%
        \ifstrequal{#1}{4}{\textcolor{teal}{R#1}}{%
                           \textcolor{cyan}{R#1}%
        }}}}%
    }%
}}

\usepackage{accv}

\usepackage{accvabbrv}

\usepackage{graphicx}
\usepackage{booktabs}

\usepackage[accsupp]{axessibility}  

\usepackage{hyperref}

\usepackage{orcidlink}

\usepackage{hyperref}
\usepackage{xr-hyper}

\makeatletter
\newcommand*{\addFileDependency}[1]{
  \typeout{(#1)}
  \@addtofilelist{#1}
  \IfFileExists{#1}{}{\typeout{No file #1.}}
}
\makeatother
\newcommand*{\myexternaldocument}[1]{
    \externaldocument{#1}
    \addFileDependency{#1.tex}
    \addFileDependency{#1.aux}
}

\myexternaldocument{_supplementary}

\begin{document}

\title{\paperTitle} 

\titlerunning{JEPA for Self-Supervised Learning of MCA}

\author{Dong-Hee Kim\inst{1} \and
Sungduk Cho\inst{1} \and
Hyeonwoo Cho\inst{1} \and
Chanmin Park\inst{1} \and
Jinyoung Kim\inst{1} \and
Won Hwa Kim\inst{2}
}

\authorrunning{D.H. Kim et al.}

\institute{VUNO Inc. \and POSTECH\\
\email{queez0405@gmail.com}, \email{wonhwa@postech.ac.kr}}

\maketitle


\begin{abstract}

    In this work, we introduce Mask-JEPA, a self-supervised learning framework tailored for mask classification architectures (MCA), to overcome the traditional constraints associated with training segmentation models. 
    Mask-JEPA combines a Joint Embedding Predictive Architecture with MCA to adeptly capture intricate semantics and precise object boundaries. 
    Our approach addresses two critical challenges in self-supervised learning: 1) extracting comprehensive representations for universal image segmentation from a pixel decoder, and 2) effectively training the transformer decoder.
    The use of the transformer decoder as a predictor within the JEPA framework allows proficient training in universal image segmentation tasks. 
    Through rigorous evaluations on datasets such as ADE20K, Cityscapes and COCO, Mask-JEPA demonstrates not only competitive results but also exceptional adaptability and robustness across various training scenarios. 
    The architecture-agnostic nature of Mask-JEPA further underscores its versatility, allowing seamless adaptation to various mask classification family.
\end{abstract}

\section{Introduction}
\label{sec:intro}

Even in the recent era of large-scale data, real-world datasets for advanced tasks, \eg, instance or panoptic segmentation, suffer from an exhaustively labor-intensive labeling process. This necessity has raised the importance of self-supervised pretraining without labels.
In computer vision, invariance-based methods~\cite{chen2020simple, grill2020bootstrap, he2020momentum} and generative methods~\cite{xie2022simmim, he2022masked, bao2021beit} have facilitated the extraction of meaningful representations from the unlabeled data for downstream tasks such as classification, object detection and even for segmentation.

%
Despite recent progress in self-supervised learning, 
many existing methods, especially for image segmentation, primarily focus on pretraining the architecture of a backbone feature extractor (\eg, ResNet~\cite{he2016deep} or ViT~\cite{dosovitskiy2020image}) \cite{chen2020simple, grill2020bootstrap, he2020momentum, xie2022simmim, he2022masked, bao2021beit}, or they are often confined to specific semantic segmentation challenges~\cite{brempong2022decoder, yang2022fully}. 
Recently, techniques based on mask classification have demonstrated the proficiency of a single architecture in addressing multiple segmentation tasks~\cite{cheng2021per, cheng2022masked, zhang2023mp, jain2023oneformer, wang2023dformer}. These tasks include semantic, instance, and panoptic segmentation, collectively referred to as universal image segmentation, all within a unified framework.
In the mask classification architectures (MCA), a \textit{pixel decoder} is used to obtain pixel-wise binary masks by utilizing features from the \textit{backbone}. 
This is followed by the \textit{transformer decoder} classifying each instance, leading to the classification of each mask.
This approach enables a universal architecture for three types of image segmentation.


\begin{wrapfigure}{r}{0.45\columnwidth}
\vspace{-15pt}
    \centering
    \includegraphics[width=1.\linewidth]{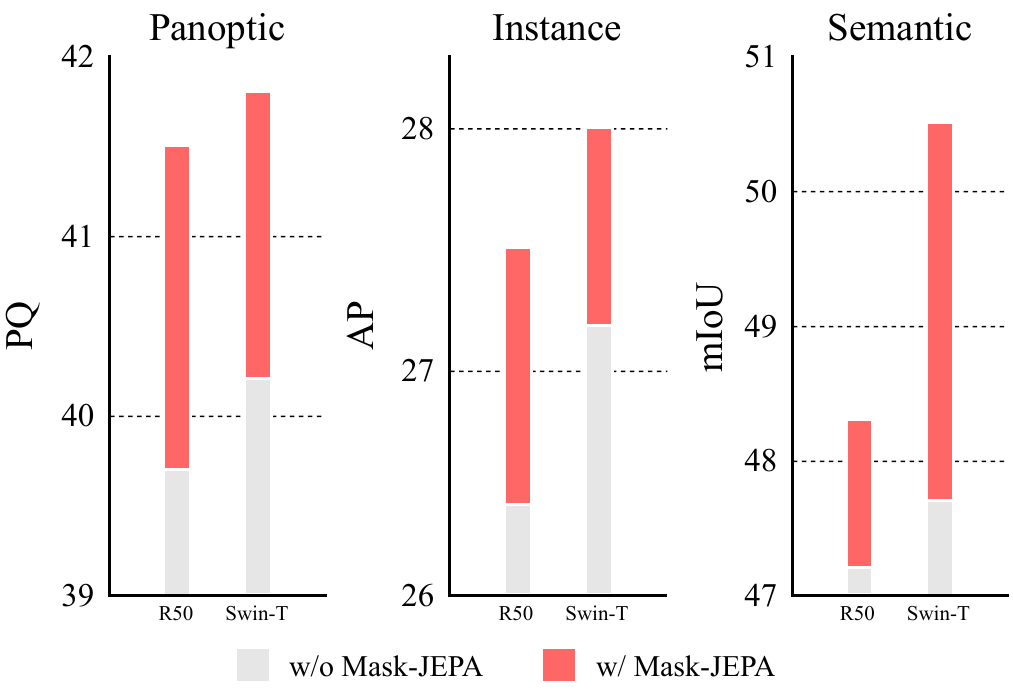}
    \vspace{-5pt}
    \caption{The graph demonstrates that the Mask-JEPA improves universal image segmentation performance across ResNet-50 and Swin Transformer Tiny backbone with Mask2Former, as evidenced by higher PQ, AP, and mIoU scores.}
    \label{fig:intro_graph}
    \vspace{-10pt}
\end{wrapfigure}
While exploring the use of self-supervised learning for entire MCAs to obtain valuable weights for rich representation, we encounter inherent limitations.
A critical component under consideration is the \textit{transformer decoder}, responsible for classifying binary masks from the pixel decoder, making it an indispensable element in MCAs. 
Consequently, while un-, semi-, and self-supervised training methods train the backbone and pixel decoder in segmentation architectures, they often lack a sophisticated mechanism to adequately train the transformer decoder. 
Therefore, self-supervised training methods for mask classification must simultaneously tackle the following two tasks: \romannumeral 1) extracting useful representations for universal image segmentation from the pixel decoder, and \romannumeral 2) properly training the transformer decoder.

To address these challenges, we present \textit{Mask-JEPA}: A Joint Embedding Predictive Architecture for Self-Supervised Learning of Mask Classification Architectures.
Observations from MCA~\cite{cheng2022masked} suggest that a pixel decoder that captures both the broad and detailed semantics of objects may significantly enhance universal image segmentation. 
In response, the Joint Embedding Predictive Architecture (JEPA)~\cite{lecun2022path} has emerged as a powerful tool. 
It efficiently extracts key semantic features without the heavy reliance on contrastive samples, overcoming the limitations of traditional invariance-based methods and offering a substantial advantage for segmentation tasks.

%
Moreover, training of the transformer decoder can be done by regarding it as a \textit{predictor} in the original JEPA framework. 
In JEPA, the role of the predictor is to forecast the feature embeddings of another encoder. 
Thus, we have adapted the transformer decoder to predict feature embeddings from the target pixel decoder. 
Furthermore, prior work~\cite{assran2023self} demonstrates that JEPA's predictor is also capable of discerning spatial and semantic embeddings. 
We empirically examine the advantages of using the transformer decoder as a predictor within the JEPA framework (See Section~\ref{subsec:ablation} discussions).

Building upon JEPA and inspired by the success of recent diffusion models~\cite{ho2020denoising, song2020denoising}, we assume that by infusing Gaussian noise into the input image and anticipating the noise through denoising, we can mimic and search for critical details (e.g., edges) from masks from MCA.

Our Mask-JEPA is applicable to any mask classification method and backbone. 
Fig.~\ref{fig:intro_graph} demonstrates the expandability of Mask-JEPA with ResNet-50~\cite{he2016deep} and Swin Transformer-Tiny~\cite{liu2021swin}. 
It is effective across various architectures, including both CNNs and ViTs, enhancing universal image segmentation performance. 
Furthermore, the architecture of most mask classification methods is predominantly built upon Mask2Former~\cite{cheng2022masked}, which implies that the weights pre-trained on Mask2Former can be directly utilized. In our experiments, we fine-tuned MP-Former~\cite{zhang2023mp} and OneFormer~\cite{jain2023oneformer}  using Mask-JEPA pretrained weights within the Mask2Former architecture, resulting in equal or enhanced performance in universal image segmentation.
This approach has demonstrated performance enhancements in various scenarios, even with limited labeled data availability.

{\bf Our contributions} are threefold:
\begin{itemize}
    \item 
    We present \textit{Mask-JEPA}, a self-supervised pretraining technique tailored for MCA 
    for universal image segmentation. 
    %
    
    \item 
    We empirically show that our methodology effectively extracts essential features for accurate image segmentation. 
    %
    
    \item 
    \textit{Mask-JEPA} shows adaptability, being architecture-agonostic for both CNNs and ViT series, through enhancements in MCAs. 
\end{itemize}
The features extracted via \textit{Mask-JEPA} demonstrate advantages in representing intricate semantics and effective edges of objects, 
and they aid instance masks to well accentuate objects, leading to better performance.
In addition, 
pretrained parameters originating from the Mask2Former~\cite{cheng2022masked} can be directly applied to the other MCAs, which removes the burden of retraining for each architectures. 

\vspace{-3pt}
\section{Preliminaries}
\label{sec:preliminary}
\vspace{-2pt}

\subsection{Mask Classification for Universal Image Segmentation}
\label{subsec:mask_cls}

Segmentation is traditionally approached as a per-pixel class probability prediction over $K$ categories. 
While per-pixel classification involves assigning a class to each individual pixel of an image, mask classification aims to categorize entire regions or masks within an image and avoid per-pixel labeling. 
Rather than labeling individual pixels, labels or categories are assigned to masks or regions of interest. 
An input RGB image, \( x \in \mathbb{R}^{3\times H \times W} \), is decomposed into $N$ instances through binary masks $\left\{m_{i}|m_{i}\in \left [ 0,1 \right ]^{H\times W} \right\}_{i=1}^{N}$ spanning $K+1$ categories. 
%
including an auxiliary "no object" label in addition to the $K$ category labels~\cite{cheng2021per}.
Predictions, which encompass both mask prediction and category probability distribution, and ground truth segments are aligned using bipartite matching, similar to the approach in DETR~\cite{carion2020end}.
Consequently, mask classification allows the prediction of multiple masks associated with a single class, rendering it apt for both semantic segmentation and instance-level segmentation tasks.

\subsection{Mask2Former Architecture}
\label{subsec:m2f_arch}

MCA~\cite{cheng2021per, cheng2022masked, zhang2023mp, jain2023oneformer, wang2023dformer} commonly consist of three foundational components: the backbone $ f_\text{back} $, the pixel decoder $ f_\text{pixel}$, and the transformer decoder $ f_\text{trans} $. In this work, our subsequent discussions will be benchmarked against the widely-adopted Mask2Former~\cite{cheng2022masked}.

The \textit{backbone} is designed to extract coarse-grained feature representations from an image. Models such as CNN~\cite{he2016deep} and ViTs~\cite{dosovitskiy2020image, liu2021swin} can be utilized as the backbone. Following this extraction, the \textit{pixel decoder} employs a pixel-wise decoding mechanism that iteratively refines and upsamples these features. This process results in high-resolution, per-pixel embeddings $ \mathcal{F}_{i} $, as discussed in~\cite{lin2017feature, zhu2020deformable}, where $ i $ denotes $ i \in \{1,2,\cdots,i_\text{last}\} $. This can be represented as: 

\begin{equation}
    \mathcal{F}_{i} = f_\text{pixel}(f_\text{back}(x)),\quad \mathcal{F}_{i} \in \mathbb{R}^{C \times \frac{H}{s_i} \times \frac{W}{s_i}}
\end{equation}

\noindent
where $s_i$ denotes scaling factor from pixel decoder.

Subsequently, the transformer decoder processes a subset of these features \( \mathcal{F}_{i} \)
along with learnable queries \( \mathcal{Q} \)~\cite{vaswani2017attention}. The outcome of this decoding process includes $ N $ masks \( \mathcal{F}_\text{mask} \in [0,1] \) and the class of each mask \( \mathcal{C} \):
\begin{equation}
    \{\mathcal{F}_\text{mask}, \mathcal{C}\} = f_\text{trans}(\mathcal{F}_{i}, \mathcal{Q})
\end{equation}
Here, \( \mathcal{Q} \) represents the \( N \) query features, each of dimension \( C \) , which help to classify the $ N $ instance masks.

\section{Method}
\label{sec:method}

\subsection{Intuition from Mask Classification}
\label{subsec:method_int}

\begin{figure*}[tp]
    \centering
    \includegraphics[width=\linewidth]{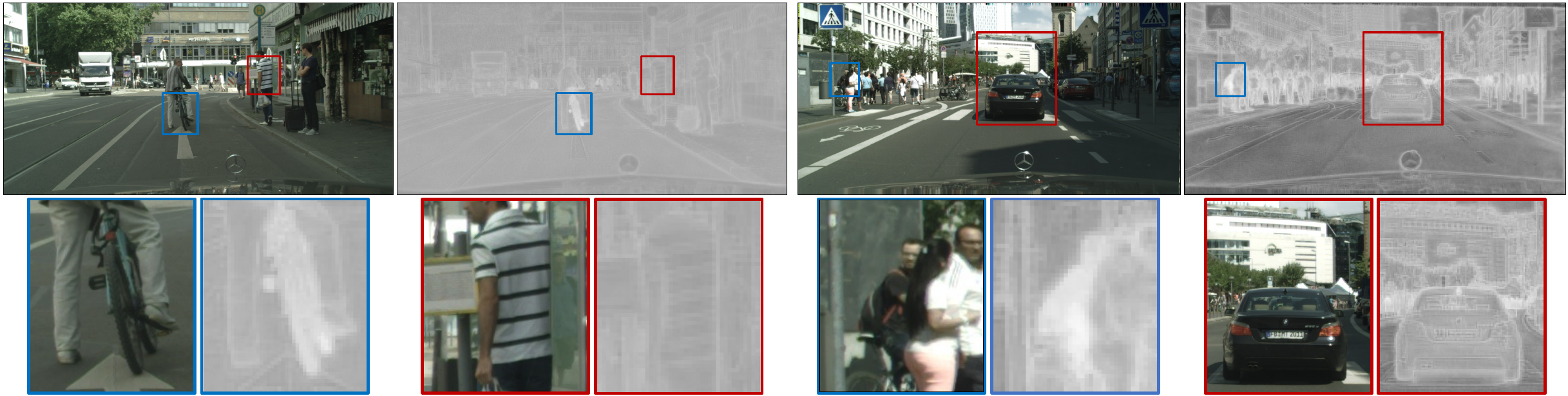}
    \vspace{-12pt}
    \caption{\textbf{Visualization of $\mathcal{F}_\text{mask}$ from a well-trained Mask2Former with segmentation labels.} A trained Mask2Former with segmentation labels not only accurately masks each object (\textcolor{blue}{blue box}) but also sharply captures edges (\textcolor{red}{red box}). The  Mask-JEPA is designed to mimic these behaviors {\em without} segmentation labels.}
    \label{fig:m2f_edge}
    \vspace{-15pt}
\end{figure*}


We first investigate what masks are extracted from MCA. In Fig.~\ref{fig:m2f_edge}, we visualize one of $ N $ mask predictions $ \mathcal{F}_\text{mask} $ from a well-trained Mask2Former. 
We found that each mask emphasizes the edges of instances, even if they are not aligned with a specific instance. This observation prompts the realization that by 1) comprehending the inherent semantics of each object and 2) effectively pinpointing edges, one could craft a robust feature extractor for universal image segmentation~\cite{chen2016semantic, li2020improving, yuan2020segfix}.

The recently proposed Joint Embedding Predictive Architecture (JEPA)~\cite{lecun2022path}, particularly its image-centric variant, Image-based JEPA (I-JEPA)~\cite{assran2023self}, has illustrated a commendable proficiency in extracting salient semantic representations. Impressively, they surpass contrastive approaches in high-dimensional feature representation efficiency~\cite{assran2023self, lecun2022path}. Such an advantage becomes pronounced in image segmentation, given that the feature sizes are typically substantial. On the other hand, many traditional contrastive methods require an plenty of contrastive samples~\cite{chen2020simple, he2020momentum, caron2020unsupervised}, posing challenges in the pretraining of segmentation architectures.

Another advantage of adopting JEPA is its reconstruction segment in I-JEPA, termed as predictor. 
It is proficient at extracting pivotal representations for feature reconstruction~\cite{assran2023self}. This capability is particularly luminous for mask classification architectures. Given the tandem operation of the pixel decoder and transformer decoder, leveraging the transformer for feature reconstruction permits the transformer decoder to assimilate representations enriched with both spatial and semantic context~\cite{bordes2022high}. Through some initial analyses, we observed that integrating the backbone and pixel decoder as JEPA's encoder, and designating the transformer decoder as the predictor, facilitates a robust representation conducive for universal image segmentation.

Denoising autoencoders, adhering to the classical approach of incorporating noise into the input and expecting to recovering the original image, prove to be particularly apt for dense prediction models. Such suitability arises from their inherent ability to be distinctly defined on a pixel-by-pixel basis~\cite{vincent2008extracting, vincent2011connection}. Recent diffusion models, capitalizing on this facet, have adeptly employed denoising, thus ensuring the retention of the minute edge details in their synthetic images (\eg, generating lifelike hair textures)~\cite{song2020denoising, ho2020denoising}. In light of these observations, we expected that such denoising autoencoders can be adeptly trained to mine salient features pivotal for mask predictions in universal image segmentation.

\subsection{Overview}
\label{subsec:method_overview}

\begin{figure*}[tp]
    \centering
    \includegraphics[width=\textwidth]{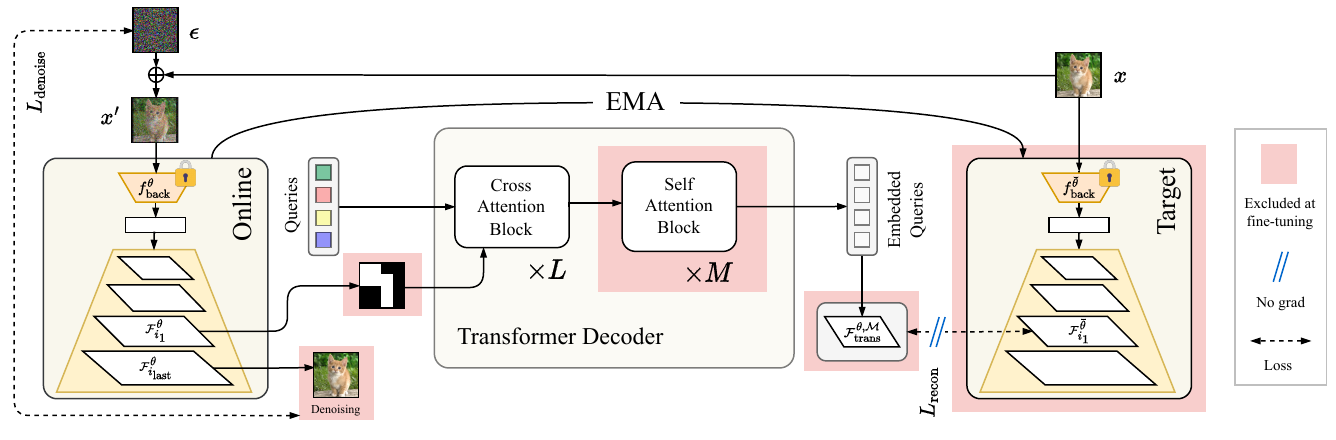}

\vspace{-5pt}    
    \caption{\textbf{Mask-JEPA Overview.} Mask-JEPA features an online mask classifier and a target backbone with a pixel decoder, updated through an exponential moving average from online versions . The target model processes image $x$, while the online model handles $x'$ with Gaussian noise $\epsilon$. The online transformer decoder processes features $\mathcal{F}_{i_1}$ from the pixel decoder along with random queries. In this step, features $\mathcal{F}_{i_1}$ are subjected to a masking process and replaced with mask tokens. The decoder's output predicts features in the target pixel decoder. Lastly, feature $\mathcal{F}_{i_\text{last}}$ from the pixel decoder undergoes a $1 \times 1$ convolution, predicting the original image $x$.}
\vspace{-10pt}
    \label{fig:overall_arch}
\end{figure*}

{
\setlength{\spaceskip}{2.9pt}
This section describes the inclusive architecture of Mask-JEPA. Our design combines an online mask classifier (consists of $ f^{\theta}_{\text{back}}$, $f^{\theta}_{\text{pixel}}$ and $f^{\theta}_{\text{trans}} $) and a target segmentation module (consists of $ f^{\bar{\theta}}_{\text{back}}$ and $ f^{\bar{\theta}}_{\text{pixel}} $). The target segmentation module is refined by employing exponential moving average (EMA) of the online segmentation module, serving as the target network, rather than utilizing fixed weights. 
}

An overview of this architecture can be seen in Fig. \ref{fig:overall_arch}. 
The input images processed by the online backbone \(f_{\text{back}}^{\theta}\) and the target backbone \(f_{\text{back}}^{\bar{\theta}}\) are distinct. To be specific, the target backbone ingests an image $x$, while the online backbone deals with the image $x' = x + \epsilon$ where $\epsilon$ denotes random Gaussian noise with zero mean and standard deviation $\sigma$. 
In this configuration, we leverage the outputs from the online pixel decoder \(f_{\text{pixel}}^{\theta}\), especially \(\mathcal{F}_{i_1}^{\theta}\) and \(\mathcal{F}_{i_{\text{\text{last}}}}^{\theta}\).

In this context, \(\mathcal{F}_{i_1}^{\theta}\) represents the feature from the online pixel decoder \(f_{\text{pixel}}^{\theta}\) and then the online transformer decoder \(f_{\text{trans}}^{\theta}\) performs the masked feature reconstruction task (Section~\ref{subsec:mfeature_recon}). On the other hand, \(\mathcal{F}_{i_{\text{last}}}^{\theta}\) is employed for Gaussian noise denoising and denotes the highest resolution feature emerging from the last layer of \(f_{\text{pixel}}^{\theta}\) (Section~\ref{subsec:denoise}). By default, we employed \(s_{i_1} = 8\) and \(s_{i_{\text{last}}} = 4\) as our settings. 

Overall, Mask-JEPA recast the backbone and pixel decoder as the encoder in JEPA, while simultaneously performing denoising of Gaussian noise. Additionally, the transformer decoder is recast as the predictor by masking image features from the pixel decoder, enabling the extraction of useful features from the transformer decoder. This approach not only succeeded in enabling MCAs to recognize semantic objects and perform edge detection but also in successfully initializing the transformer decoder with improved weights (Section~\ref{subsec:ablation}).

\subsection{Masked Feature Reconstruction}
\label{subsec:mfeature_recon}

In Mask-JEPA, the masked feature gets replaced with mask tokens confined to the spatial resolution of $ \mathcal{F}_{i_1} $ by random patch sampling, aligning with methods such as MAE \cite{he2022masked} and SimMIM \cite{xie2022simmim} but with slight differences. 

Given the feature $ \mathcal{F}_{i_1}^{\theta} $ with shape $ C \times H \times W $, we segment $ H \times W$ into tiled patches of size $ p \times p $. Each patch belongs to one of the two states: 1) fully visible or 2) fully masked. 
Masked portions are uniformly substituted with a learnable mask token with a length of $ C $. 
The decision to fully expose or mask a patch is based on a preset masking ratio.

This masking process is repeated over \( L \) iterations of the cross-attention blocks, each consisting of cross-attention, self-attention~\cite{vaswani2017attention}, and a feedforward network. The most characteristic feature of the cross-attention block is the cross-attention mechanism, originating from Mask2Former~\cite{cheng2022masked}, which is defined as:
\begin{equation}
    \mathbf{X}_{l} = \text{softmax}(\mathbf{Q}_{l}\mathbf{K}_{l}^{T})\mathbf{V}_{l} + \mathbf{X}_{l-1}
\end{equation}
where \( l \) denotes the layer index, and \( \mathbf{X}_l \) represents the \( N \) query features, each of dimension \( C \), at layer \( l \). 
\( \mathbf{Q}_l \) is defined as \( f_Q(\mathbf{X}_{l-1}) \) via linear function \( f_Q (\cdot)\). The initial query features fed to the transformer decoder are represented by \( \mathbf{X}_0 = \mathcal{Q} \). The masked image features \( \mathbf{K}_l, \mathbf{V}_l \in \mathbb{R}^{H_{i_1}W_{i_1} \times C} \) undergo transformations via linear functions \( f_K(\cdot) \) and \( f_V(\cdot) \), respectively, and $H_{i_1}$ and $W_{i_1}$ are the spatial resolution of image features of $\mathcal{F}_{i_1}$. Additionally, a sinusoidal positional embedding $e_{\text{pos}} \in \mathbb{R}^{H_{i_1}W_{i_1} \times C}$ is added at $\mathbf{K}_l, \mathbf{V}_l$ following~\cite{carion2020end}. The final output, \( \mathbf{X}_L \), is fed into linear function \( f_L(\cdot) \) to reconstruct the target pixel decoder's embedding feature, \( F_{i_1}^{\bar{\theta}} \).

A distinct feature of Mask-JEPA is the introduction of \( M \) additional self-attention blocks~\cite{vaswani2017attention} which enables richer feature extraction~\cite{park2022self}. These are added orthogonally to the cross-attention blocks. This relationship is expressed as: 
\begin{equation}
    \mathcal{F}_{\text{trans}}^{\theta, \mathcal{M}} = f_{\text{trans}}^{\theta}(\mathcal{M}(\mathcal{F}_{i_1}^{\theta}), \mathcal{Q})
\end{equation}
where \( \mathcal{M} \) denotes the masking operation, and \( f_{\text{trans}}^{\theta} \) is the adapted transformer decoder tasked with reconstructing the feature embeddings. 
The refined embeddings are then compared with the layer normalized~\cite{ba2016layer} target pixel decoder's output \( \mathcal{F}^{\bar{\theta}}_{i_1} \):
\begin{align}
    L_{\text{recon}} &= D_{\mathcal{M}}(\mathcal{F}_{\text{trans}}^{\theta, \mathcal{M}}, \text{LN}(\mathcal{F}_{i_1}^{\bar{\theta}})).
    \label{eq:L_recon}
\end{align}
In this context, \( D_{\mathcal{M}} \) measures the distance between two features, computed only within regions masked by \( \mathcal{M} \). We adopted $\ell_2$-distance for \eqref{eq:L_recon}.

Overall, masked feature reconstruction can be characterized as concurrently performing dual denoising tasks: one via direct mask reconstruction facilitated by the transformer decoder's masking operation, and the other through predicting features derived from both the original image $ x $ and its perturbed version $ x' $.

\subsection{Gaussian Noise Denoising}
\label{subsec:denoise}

Denoising schemes provide fine details at a per-pixel level and preserve sharp edges. As previously mentioned, $ \mathcal{F}_{i_{\text{last}}}^{\theta} $ is used for Gaussian noise denoising. 
Due to the size mismatch between the input image $ x $ and the dimensions of $ \mathcal{F}_{i_{\text{last}}}^{\theta} $, 
a direct noise prediction matching the size of $ x $ is impractical. 
To address this, 
we produce Gaussian noise of dimensions $ H/s_{i_{\text{last}}} $ and $ W/s_{i_{\text{last}}} $ and upscale it by a factor of $s_{i_{\text{last}}}$. Rather than tiling $ H/s_{i_{\text{last}}} $ and $ W/s_{i_{\text{last}}} $ sized Gaussian noise in a $ s_{i_{\text{last}}} \times s_{i_{\text{last}}} $ pattern, we choose to expand each pixel to occupy a $ s_{i_{\text{last}}} \times s_{i_{\text{last}}} $ area. The subsequent denoising step involves a $ 1 \times 1 $ convolution, resulting in an output of size $ 3 \times H/s_{i_{\text{last}}} \times W/s_{i_{\text{last}}} $. 
The loss formulation for the denoising operation is:
\begin{equation}
    L_{\text{denoise}} = D(\text{Conv}(\mathcal{F}_{i_{\text{last}}}^{\theta}), x). 
\end{equation}
Similarly to \eqref{eq:L_recon}, \( D \) represents the  distance between two features equipped with $\ell_2$-distance by default.  
The success of diffusion models~\cite{song2020denoising, ho2020denoising} in predicting $ \epsilon $ instead of raw image $ x $ offers another option for denoising. 
We discuss predicting $ x $ or $ \epsilon $ in detail in the Appendix~\ref{supp:denoising_target}. 

Consequently, our final loss function given as:
\begin{equation}
    L_{\text{final}} = L_{\text{recon}} + L_{\text{denoise}}
\end{equation}
which combines reconstruction and denoising losses. 

\section{Experiments}
\label{sec:results}

We start by detailing the experimental setup and datasets. Then, we showcase Mask-JEPA performance in various image segmentation tasks, including finetuning under different data conditions. 
At the end, we demonstrate ablations and behavior analysis of Mask-JEPA.

\subsection{Implementation}
\label{subsec:implement}

\textbf{Model Training:} During the pretraining phase, the Gaussian noise intended for denoising has a standard deviation of $\sigma = 0.4$. Additionally, a masking ratio of 0.5 and a patch size $p$ of 8 are employed for feature reconstruction. We set the number of cross-attention blocks \( L \) and self-attention blocks \( M \) to 9 and 2, respectively. The image resolution \( H \times W \) is set to \( 512 \times 512 \).



In the fine-tuning phase, our settings align with those of prevalent mask classification methods, such as Mask2Former~\cite{cheng2022masked}, MP-Former~\cite{zhang2023mp}, and OneFormer~\cite{jain2023oneformer}.  We evaluate the efficacy of our approach using the ResNet50~\cite{he2016deep} and Swin-Transforemr Tiny~\cite{liu2021swin} architectures as a backbone. More extensive details are provided in Appendix \ref{supp:imp_detail}. 

\nbf{Datasets and evaluation: } We used ImageNet ILSRVC 2012 (IN1K) dataset~\cite{russakovsky2015imagenet} which contains 1.2M images for self-supervised pretraining. We ran Mask-JEPA on rigorous testing on three datasets: Cityscapes~\cite{cordts2015cityscapes, cordts2016cityscapes}, ADE20K~\cite{zhou2019semantic} across all three universal image segmentation, and COCO 2017~\cite{lin2014microsoft} specifically for panoptic and instance segmentation tasks. 
For the instance segmentation, performance was measured using the mask AP~\cite{lin2014microsoft} for instances labeled as ``thing'' within the images. 
Semantic segmentation performance is determined through mean Intersection-over-Union (mIOU) across all classes, inclusive of foreground and background. 
For panoptic segmentation, we employ the panoptic quality (PQ) metric~\cite{kirillov2019panoptic} for assessment.

\subsection{Universal Image Segmentation in Full Data Regimes}
\label{subsec:fUll_data}



\begin{table*}[t]
    \tablestyle{4pt}{1.2}\scriptsize
    {\fontfamily{ptm}\selectfont
    \resizebox{\textwidth}{!}{
    \begin{tabular} {c|l | x{25}x{25}x{25} | x{18}x{18}x{18}x{18} |x{36}x{36}}
    \hline
        \multirow{2}{*}{Method} & \multirow{2}{*}{Backbone} & \multicolumn{3}{c|}{Panoptic } & \multicolumn{4}{c|}{Instance } & Semantic  \\
        &  & PQ & AP$^\text{Th}_\text{pan}$ & mIoU$_\text{pan}$ & AP & AP$^\text{S}$ & AP$^\text{M}$ & AP$^\text{L}$ & mIoU \\
        \shline
        MaskFormer~\cite{cheng2021per} &R50 & 34.7\phantom{$^*$} & - & - & - & - & - & - & -  \\
        Mask2Former~\cite{cheng2022masked} & R50\phantom{$^{\text{\textdagger}}$}
        & 39.7\phantom{$^*$} & 26.5 & 46.1\phantom{$^*$} & 26.4 & 10.4 & 28.9 & 43.1 & 47.2/46.6$^*$ \\
        \textbf{\quad+Mask-JEPA} & R50\phantom{$^{\text{\textdagger}}$}
        & \textbf{41.5\dt{+1.8}} & \textbf{27.5} & \textbf{47.0} & \textbf{27.5\dt{+1.1}} & \textbf{11.3} & \textbf{29.5} & \textbf{45.3} & \textbf{48.3\dt{+1.7}} \\
        \hdashline
        
        MP-Former~\cite{zhang2023mp} & R50\phantom{$^{\text{\textdagger}}$}
        & 40.8 & 27.1 & \textbf{48.3} & \textbf{28.0} & 10.5 & \textbf{30.7} & 44.6 & \textbf{48.1} \\
        \textbf{\quad+Mask-JEPA} & R50\phantom{$^{\text{\textdagger}}$}
        & \textbf{41.4\dt{+0.6}} & \textbf{28.4} & 47.8 & \textbf{28.0} & \textbf{11.5} & \textbf{30.6} & \textbf{45.2} & \textbf{48.1} \\
        \hdashline
        
        OneFormer~\cite{jain2023oneformer} & R50\phantom{$^{\text{\textdagger}}$}
        & 41.9 & 27.3 & 47.3 & - & - & - & - & - \\
        \textbf{\quad+Mask-JEPA} & R50\phantom{$^{\text{\textdagger}}$}
        & \textbf{42.4\dt{+0.5}} & \textbf{27.7} & \textbf{47.4} & - & - & - & - & - \\
        \hline
        Mask2Former~\cite{cheng2022masked} & Swin-T\phantom{$^{\text{\textdagger}}$}
        & 40.2$^*$ & 27.1$^*$ & 48.6$^*$ & 27.2$^*$ & 10.2$^*$ & 29.7$^*$ & 45.9$^*$ & 47.7 \\
        \textbf{\quad+Mask-JEPA} & Swin-T\phantom{$^{\text{\textdagger}}$}
        & \textbf{41.8\dt{+1.6}} & \textbf{28.0} & \textbf{49.7} & \textbf{28.5\dt{+1.3}} & \textbf{11.4} & \textbf{31.0} & \textbf{47.1} & \textbf{50.5\dt{+2.8}} \\
        \hdashline

        MP-Former~\cite{zhang2023mp} & Swin-T\phantom{$^{\text{\textdagger}}$}
        & 41.5$^*$ & 28.0$^*$ & 48.5$^*$ & 28.3$^*$ & 10.6$^*$ & 30.6$^*$ & 47.6$^*$ & 48.6$^*$ \\
        \textbf{\quad+Mask-JEPA} & Swin-T\phantom{$^{\text{\textdagger}}$}
        & \textbf{42.6\dt{+1.1}} & \textbf{28.6} & \textbf{50.7} & \textbf{29.4\dt{+1.1}} & \textbf{11.1} & \textbf{32.0} & \textbf{48.5} & \textbf{49.5\dt{+0.9}} \\
        \hdashline
        
        OneFormer~\cite{jain2023oneformer} & Swin-T\phantom{$^{\text{\textdagger}}$}
        & 42.7$^*$ & 28.6$^*$ & 49.3$^*$ & - & - & - & - & - \\
        \textbf{\quad+Mask-JEPA} & Swin-T\phantom{$^{\text{\textdagger}}$}
        &\textbf{44.2\dt{+1.5}} & \textbf{29.7\dt{+1.1}} & \textbf{50.1\dt{+0.8}} & - & - & - & - & - \\
        \hline
    
    \end{tabular}
    }}
    \caption{\textbf{Image segmentation results on ADE20K \texttt{val} in full data regimes.}  Mask-JEPA improves the performance on all three segmentation tasks with R$50$ and Swin-T as backbone. All metrics are evaluated with \emph{single-scale} inference. 
    `-': the results are not reported nor implementation not reproducible.  
    $^*$: Results reproduced using the official code. }    
\vspace{-15pt}
\label{tab:benchmark:ade20k}

\end{table*}
\begin{table*}[h]
    \centering
    {\fontfamily{ptm}\selectfont
    \resizebox{\textwidth}{!}{
    \begin{tabular}{l|l |l | x{28}x{28}x{28} | x{28}x{28} |x{36}x{36}}
        \hline
        \multirow{2}{*}{Method} & \multirow{2}{*}{Backbone} & \multirow{2}{*}{Segmentor} & \multicolumn{3}{c|}{Panoptic } & \multicolumn{2}{c|}{Instance } & {Semantic } \\
        &  & & PQ & AP$^\text{Th}_\text{pan}$ & mIoU$_\text{pan}$ & AP & AP50 & mIoU\\
        \shline
        Segsort~\cite{hwang2019segsort} & R50 & PSPNet~\cite{zhao2017pyramid} & - & - & - & - & - & 78.2  \\
        PC$^2$Seg~\cite{zhong2021pixel} & R50 & DeepLabV3+~\cite{chen2018encoder} & - & - & - & - & - & 75.4  \\
        AuxContrast~\cite{zhang2021looking} & R50 & DeepLabV3+~\cite{chen2018encoder} & - & - & - & - & - & 79.6  \\
        SlotCon~\cite{wen2022self} & R50 & FPN~\cite{lin2017feature} & - & - & - & - & - & 76.3  \\
        DDeP$^{\text{\textdagger}}$~\cite{wu2022denoising} & R50 & TransUNet~\cite{chen2021transunet} & - & - & - & - & - & 80.6  \\
        No Pretrain & R50 & Panoptic-DeepLab~\cite{cheng2020panoptic} & 60.3  & 32.1 & \textbf{78.7} & - & - & -  \\
        \hdashline
        No Pretrain & R50 & Mask2Former~\cite{cheng2022masked} & \textbf{62.1}  & 37.3 & 77.5 & 37.4 & 61.9 & 79.4  \\
        \textbf{Mask-JEPA (Ours)} & R50 & Mask2Former~\cite{cheng2022masked} & \textbf{62.2}  & \textbf{38.1} & 78.5 & \textbf{37.9} & \textbf{62.4} & \textbf{80.7\dt{+1.3}}  \\
        \hline
        No Pretrain & Swin-T & Mask2Former~\cite{cheng2022masked} & 63.4$^*$  & 38.7$^*$ & \textbf{80.9}$^*$ & 39.0$^*$ & 65.8$^*$ & 81.3$^*$ \\
        \textbf{Mask-JEPA (Ours)} & Swin-T & Mask2Former~\cite{cheng2022masked}  & \textbf{64.4\dt{+1.0}}  & \textbf{39.4} & 80.5 & \textbf{39.3} & \textbf{66.1} & \textbf{82.3\dt{+1.0}} \\
        \hline
    \end{tabular}
    }}
    \caption{\textbf{Image segmentation results on Cityscapes \texttt{val} in full data regimes.} We report results on all three segmentation tasks with R50 or Swin-T as backbone. All metrics are evaluated with \emph{single-scale} inference. 
    $^{\text{\textdagger}}$: Backbone + Segmentor pre-trained on ImageNet-22K. 
    $^*$: Results reproduced using the official code. }
    \vspace{-10pt}
    \label{tab:benchmark:cityscapes_full}
\end{table*}


\begin{table}[!h]
    \centering
    \tablestyle{3pt}{1.2}\scriptsize
    \resizebox{0.7\columnwidth}{!}{    
    \begin{tabular} {l|c | x{28}x{28}x{28} | x{15}x{15}x{15}x{15} }
    \hline
        \multirow{2}{*}{Method} & \multirow{2}{*}{Backbone} & \multicolumn{3}{c|}{Panoptic } & \multicolumn{4}{c}{Instance } \\        
        &  & PQ & AP$^\text{Th}_\text{pan}$ & mIoU$_\text{pan}$ & AP & AP$^\text{S}$ & AP$^\text{M}$ & AP$^\text{L}$ \\
        \shline
        MaskFormer~\cite{cheng2021per} & R50 & 46.5 & 33.0 & 57.8 & 34.0 & 16.4 & 37.8 & 54.2  \\
        Mask2Former~\cite{cheng2022masked} & R50
        & 51.4$^*$ & 41.9 & 61.5 & \textbf{43.7} & \textbf{23.4} & \textbf{47.2} & 64.8 \\
        \quad\textbf{+ Mask-JEPA} & R50
        & \textbf{52.0} & \textbf{42.1} & \textbf{61.7} & 43.6 & 22.9 & 47.0 & \textbf{65.3} \\
        
        \hline
    \end{tabular}
    }
    \caption{\textbf{Image segmentation on COCO \texttt{val2017} in full data regimes.}  Panoptic segmentation has 133 categories and instance segmentation has 80 categories. Mask-JEPA improves the performance on all three segmentation tasks with R$50$ as backbone. 
    All metrics are evaluated with \emph{single-scale} inference. 
    $^*$: Results reproduced using the official code. }
\vspace{-5pt}
\label{tab:benchmark:coco}
\vspace{-0.2in}
\end{table}
\begin{figure*}[tp]
    \centering
    \includegraphics[width=1.\linewidth]{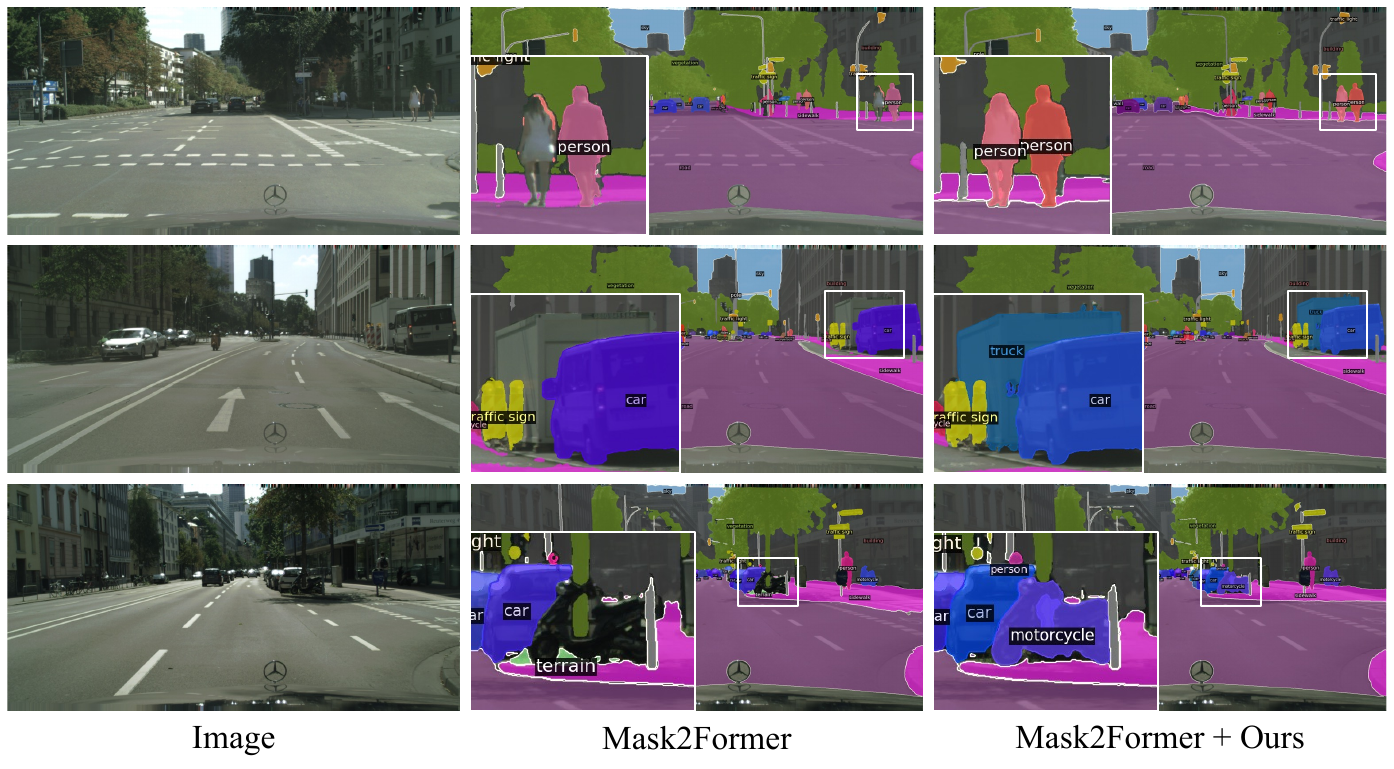}

    \caption{\textbf{Qualitative results.} Our Mask-JEPA pretrained model achieves more accurate detection and segmentation compared to plain Mask2Former training (as shown within the white boxes). Both trained with Swin-T backbone. Zoom in for a closer view.}
    \vspace{-10pt}
    \label{fig:output}
\end{figure*}

We investigate the performance of Mask-JEPA in the context of fully annotated labels for fine-tuning.

\noindent
\textbf{Pretraining:} 
%
We used the pretrained backbone weights from IN1K classification and kept them frozen. Then, we pretrained Mask-JEPA on the unlabeled IN1K for 5 epochs.

\noindent
\textbf{Baselines:}
We compared our Mask-JEPA to state-of-the-art un-, semi-, and self-supervised segmentation methods on the Cityscapes dataset. 
For unsupervised methods, we compared the results with transfer learning for a fair comparison. 
As these methods are focused on the semantic segmentation, 
we also compared our method with the panoptic segmentation architecture baseline~\cite{cheng2020panoptic}, 
and we kept the backbone consistent as ResNet50 across the experiments for fair comparisons. 

\noindent
\textbf{Setup:}
To evaluate Mask-JEPA, we fine-tuned it on three distinct datasets: ADE20K~\cite{zhou2019semantic}, Cityscapes~\cite{cordts2015cityscapes, cordts2016cityscapes} and MS COCO~\cite{lin2014microsoft}.
Not only did we compare with Mask2Former, but we also examined the effectiveness of Mask-JEPA pretrained weights when integrated with MP-Former and OneFormer, both of which are built upon the Mask2Former. 
For MP-Former and OneFormer, we directly adopted the pretrained weights from Mask2Former without any further training.

\noindent
\textbf{Results:} 
Table~\ref{tab:benchmark:ade20k} highlights that the models pretrained via Mask-JEPA consistently outperforms without pretraining, surpassing through all Mask2Former, MP-Former and OneFormer. 
Notably, the mIoU score achieves a 1.7-point increase compared to the baseline training using ResNet50 backbone Mask2Former.


An interesting observation is that the model trained with the Swin-T backbone is more harmonized with Mask-JEPA. We interpret this phenomenon as a result of the differences in structural capacity between ResNet and Swin-Transformer. To support this, 
we observed that Swin-T exhibits a lower loss function value compared to ResNet50. 

In Table~\ref{tab:benchmark:cityscapes_full}, the model with Mask-JEPA pretrained weights outperforms previous works utilizing unlabeled data with cityscapes dataset. DDeP~\cite{wu2022denoising} employs a ResNet backbone pretrained on the ImageNet-22K dataset, while all other methods utilize ImageNet-1K. Additionally, Mask-JEPA is not tethered to one dataset but is proficient at gleaning generalized features.
A case in point is the uplift of 1.3 mIoU points observed with the ResNet50 backbone on Cityscapes.

In Figure~\ref{fig:output}, we compare the qualitative results of Mask-JEPA and plain training. We observe that Mask-JEPA consistently refines the errors of models without pretraining. The results for self-supervised pretrained backbone weights only and for finetuning on ADE20K are available in Appendix~\ref{supp:backbone_ssl}.

\subsection{Universal Image Segmentation in Low-Data Regimes}
\label{subsec:low_data}






\begin{table}[t!]\setlength{\tabcolsep}{10pt}
  \centering
  \resizebox{.63\linewidth}{!}{
    \begin{tabular}{c |l | ccc}
    \hline

        Portion & Method & PQ & AP & mIoU \\
        \midrule
        
        \multirow{2}{*}{10\%} &  Mask2Former & 23.3 & 13.1 & 31.1 \\


        & \quad +Mask-JEPA & 25.9\textcolor{red}{\dt{+2.6}} & 14.5\textcolor{red}{\dt{+1.4}} & 31.7\textcolor{red}{\dt{+0.6}} \\

        \hdashline

        \multirow{2}{*}{5\%} &  Mask2Former & 19.0 & 9.6 & 25.2 \\


        & \quad +Mask-JEPA & 20.7\textcolor{red}{\dt{+1.7}} & 10.7\textcolor{red}{\dt{+1.1}} & 24.9\textcolor{blue}{\dt{-0.3}} \\

        \hdashline

        \multirow{2}{*}{2\%} &  Mask2Former & 11.3 & 6.2 & 17.6 \\

        & \quad +Mask-JEPA & 13.5\textcolor{red}{\dt{+2.1}} & 6.7\textcolor{red}{\dt{+0.5}} & 16.5\textcolor{blue}{\dt{-1.1}} \\

        \hdashline

        \multirow{2}{*}{1\%} &  Mask2Former & 6.9 & 3.8 & 12.5 \\

        & \quad +Mask-JEPA & 7.5\textcolor{red}{\dt{+0.6}} & 4.4\textcolor{red}{\dt{+0.6}} & 11.2\textcolor{blue}{\dt{-1.3}} \\
        \hline
    \end{tabular}
    }
    \caption{\textbf{Performance using $k\%$ labeled data on ADE20K \texttt{train}.} The models trained with Mask-JEPA outperform not-pretrained methods, especially in panoptic and instance segmentation.}
    \label{tab:low_regime}
\vspace{-10pt}
\end{table}

We investigate the performance of Mask-JEPA in the context of a small amount of annotated data for fine-tuning with ResNet50 backbone.

\noindent
\textbf{Pretraining:} We conduct the same training procedure as described in Section~\ref{subsec:fUll_data}. 

\noindent
\textbf{Setup:} We evaluated whether the representations trained with Mask-JEPA remain effective when transferred in limited data scenarios. Using randomly selected $k\%$, $k \in \{1,2,5,10\}$, of labeled data from the ADE20K \texttt{train} dataset for the semantic segmentation, we conducted evaluations on the ADE20K \texttt{val}. 
For fair comparison, we fixed the labeled images at the same proportion $k$.

\noindent
\textbf{Results:} Table~\ref{tab:low_regime} shows that Mask-JEPA mostly outperforms without pretraining methods on low-data regimes. Especially, panoptic and instance segmentation performance is consistently better than baseline.

\subsection{Ablations and Discussions}
\label{subsec:ablation}

This section explores the reasoning and factors influencing the architecture and algorithms used in the main experiments.

\vspace{5pt}
\nbf{4.4.1 Design Ablations} \\
\vspace{-5pt}

\begin{table}[t]
    \begin{minipage}[t]{0.55\textwidth}
        \centering
        \begin{minipage}[t]{0.45\columnwidth}
            \renewcommand*{\arraystretch}{1}
            \centering
            \scalebox{0.75}{
                \begin{tabular}{lc}
                \hline
                Component Ablation   & mIoU \\ \hline\toprule
                Mask-JEPA (Ours)    & \textbf{48.3}  \\
                \hline
                \quad-- $L_\text{recon}$   & 47.1\dt{-1.2}    \\ 
                \quad-- $L_\text{denoise}$ & 47.3\dt{-1.0}  \\
                \quad-- Self-Attention  & 46.9\dt{-1.4}   \\
                Mask2Former & 46.6\\
                \hline
                \end{tabular}
            }
        \end{minipage}\hspace{3pt}
        \begin{minipage}[t]{0.25\columnwidth}
            \renewcommand*{\arraystretch}{1.31}
            \centering
            \scalebox{0.86}{
            \begin{tabular}{cc}
            \hline
            \textbf{$s_i$}   &  mIoU \\ \hline \toprule
            8  & \textbf{48.3} \\ 
            16  & 47.7 \\ 
            32   & 47.6 \\ 
            \hline
            \end{tabular}}
        \end{minipage}\hspace{-14pt}
        \begin{minipage}[t]{0.25\columnwidth}
            \renewcommand*{\arraystretch}{1.18}
            \centering
            \scalebox{0.77}{
            \begin{tabular}{cc}
            \hline
            \textbf{$M$}         & mIoU \\ \hline\toprule
            0      & 46.9 \\ 
            1           & 47.6 \\
            2           & \textbf{48.3} \\
            3           & 47.6 \\
            \hline
            \end{tabular}}
        \end{minipage}
        \caption{Ablation Study on Component of Mask-JEPA. Left: Ablation on the model components. By excluding each component of Mask-JEPA, we demonstrate its effectiveness. Middle: Ablation on the resolution of $\mathcal{F}_i$($s_i$), Right: Ablation on the number of self-attention blocks.}
        \label{tab:ablation}
    \end{minipage}
    \hspace{1.mm}
    \begin{minipage}[t]{0.42\textwidth}
        \centering
        \begin{minipage}[t]{0.2\textwidth}
            \renewcommand*{\arraystretch}{1.1}
            \centering
            \scalebox{0.77}{
            \begin{tabular}{cc}
                \hline
                \textbf{$\sigma$}   &  mIoU \\ \hline \toprule
                0.2  &  47.3 \\ 
                0.3  &  47.4\\ 
                0.4  & \textbf{48.3} \\ 
                0.5  &  47.3 \\ 
                \hline
            \end{tabular}}
        \end{minipage}\hspace{3pt}
        \begin{minipage}[t]{0.2\textwidth}
            \centering
            \renewcommand*{\arraystretch}{1.37}
            \scalebox{0.77}{
                \begin{tabular}{lc}
                    \hline
                    Masking Ratio   & mIoU \\ \hline\toprule
                    0.25  &  47.3   \\ 
                    0.5   & \textbf{48.3}  \\
                    0.75  & 47.2   \\
                    \hline        
                \end{tabular}}
        \end{minipage}\hspace{40pt}
        \begin{minipage}[t]{0.2\textwidth}
            \renewcommand*{\arraystretch}{1.37}
            \centering
            \scalebox{0.77}{
            \begin{tabular}{cc}
                \hline
                \textbf{$p$}  & mIoU \\ \hline\toprule
                4      &  47.3 \\ 
                8      & \textbf{48.3} \\
                16     &  48.1\\
                \hline
            \end{tabular}}
        \end{minipage}

    \caption{Hyperparameter Robustness on Mask-JEPA. Left: Ablation on the Gaussian noise's standard deviation $\sigma$. Middle: Ablation on the masking ratio of $\mathcal{F}_{i_1}^{\theta}$, Right: Ablation on the masking patch size $p$.}
    \label{tab:hparam_ablation}

    \end{minipage}
    \vspace{-20pt}
\end{table}

\nbf{The effect of each component:} Table~\ref{tab:ablation} (left) presents the effect of each component of \textit{Mask-JEPA} on semantic segmentation in self-supervised learning. All models were initially trained on the ImageNet-1K dataset for 5 epochs, followed by fine-tuning on the ADE20K dataset with 160k iterations. 
We initially hypothesized that the key components of Mask-JEPA — its JEPA training, denoising, and extra self-attention blocks — play a significant role in extracting better representations. 
The Table \ref{tab:ablation} (left) shows 
that all the incorporated components collectively contribute to enhanced semantic segmentation performance.

\nbf{Resolution of Reconstructed $\mathcal{F}_i$:} 
We explore the impact of the resolution of the feature we aim to reconstruct. Our hypothesis posits that reconstructing the largest (and latest) feature using JEPA can enhance performance, a finding our results support. Among the resolution options $\{8, 16, 32\}$ for $s_{i}$, choosing 8 yielded the best outcomes, as demonstrated in Table~\ref{tab:ablation} (middle).

\noindent
\textbf{Number of addtional $M$ self-attention block:} Masking methods in the ViTs, especially focusing on the early layers of the model, motivate us to add an extra self-attention block. However, it is essential to determine the optimal number of extra self-attention blocks. Our empirical results in Table~\ref{tab:ablation} (right) show that $M=2$ was optimal in our settings.


\vspace{5pt}
\nbf{4.4.2 Hyperparameter Robustness} \\
\vspace{-5pt}

\nbf{Gaussian noise standard deviation $\sigma$:} Table~\ref{tab:hparam_ablation} (left) shows the ablation study on the standard deviation of Gaussian noise $\sigma$ for the denoising task in the online mask classifier. Conclusively, we find that a value of 0.4 for the standard deviation $\sigma$ is optimal in ADE20K semantic segmentation dataset.

\nbf{Masking ratio:} We examine the influence of the masking ratio on the input $\mathcal{F}_{i_1}^{\theta}$ for transformer decoders, as presented in Table~\ref{tab:hparam_ablation} (middle). Our experiments demonstrate that a masking ratio of 0.5 results in the highest mIoU score for ADE20K semantic segmentation.

\nbf{Masking patch size $p$:} The masking patch size $p$ is another important hyperparameter. The detailed results, presented in Table~\ref{tab:hparam_ablation} (right), show that a patch size of $p=8$ yields the best performance for the ADE20K semantic segmentation.

Overall, our model consistently outperforms the benchmarks, achieving higher mIoU scores than both the score we reproduced (46.6 mIoU) and the score reported in prior work (46.1 mIoU)~\cite{li2023mask}, regardless of the chosen hyperparameters.

\vspace{5pt}
\nbf{4.4.3 Discussions on the Mask-JEPA} \\
\vspace{-5pt}

\begin{figure*}[!t]
    \centering
    \includegraphics[width=\linewidth]{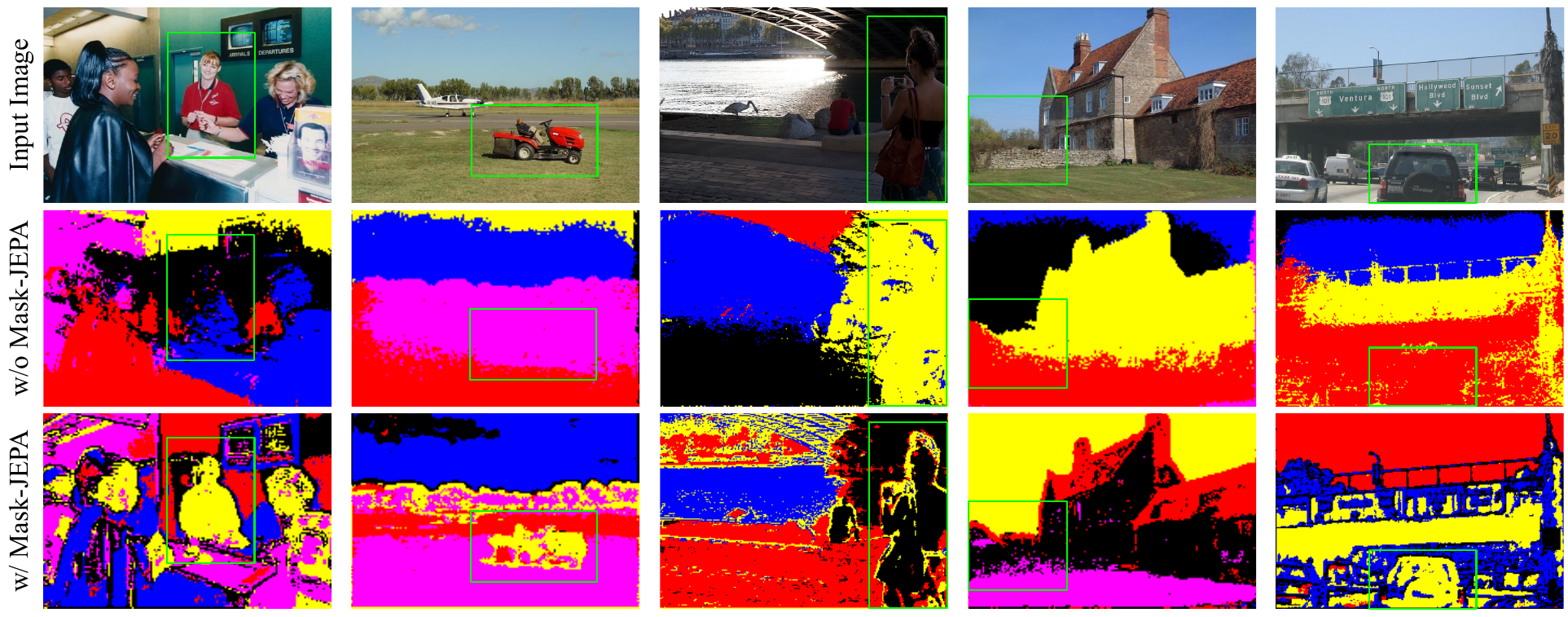}
    \vspace{-20pt}
    \caption{\textbf{Visualization of Mask-JEPA pretrained pixel decoder output.} We visualized the output $\mathcal{F}_{i_\text{last}}$ from the pixel decoder using k-means clustering. The results show that models trained with Mask-JEPA (row 3) effectively identify both semantic objects and edges. In contrast, a pixel decoder without pretraining (row 2) struggles to cluster similar semantics. \eg, in column 2, it completely fails to detect automobiles.}
    \label{fig:cluster}
    \vspace{-10pt}
\end{figure*}

\noindent
\textbf{What does Mask-JEPA capture?} We compared the $\mathcal{F}_{i_{\text{last}}}$ originating from the randomly initialized pixel decoder before Mask-JEPA pretraining with the $\mathcal{F}_{i_{\text{last}}}$ post Mask-JEPA pretraining. 
We conducted unsupervised segmentation of $\mathcal{F}_{i_{\text{last}}}$ using k-means clustering with $k=5$, whose results can be seen in Figure~\ref{fig:cluster}. 
While the unpretrained model struggles to identify objects or discern precise contours, the model pretrained with Mask-JEPA successfully detects both semantic objects and edges in the image, as we discussed in Section~\ref{subsec:method_int}. 
For instance, in Figure~\ref{fig:cluster} column 1, the person in the center is barely detectable using random initialization, whereas our method not only identifies the edges but also provides an almost accurate semantic mask. 
Furthermore, we compared the visualization of unsupervised semantic segmentation in the Appendix~\ref{supp:extended_qual}.

\noindent
\textbf{Does pretrained Mask-JEPA transformer decoder carry useful features?} While the pixel decoder's features $\mathcal{F}_{i_{\text{last}}}$, can be visualized, the transformer decoder directly predicts the feature. Therefore, instead of directly visualizing its role, we investigated its function by fine-tuning on the ADE20K dataset, excluding the specific transformer decoder weight. The results, shown in Table~\ref{tab:no_trans_decoder}, demonstrate improvements in all metrics used to evaluate universal image segmentation, indicating that the features extracted by the transformer decoder also play a crucial role in enhancing performance.

\begin{table}[t]
    \begin{minipage}[t]{0.46\textwidth}
        \centering
        \resizebox{1.\linewidth}{!}{
        \begin{tabular}{l | ccc}
        
        Setup & PQ & AP & mIoU \\
        \midrule        
        w/o Transformer Decoder & 40.2 & 27.4 & 47.3 \\
        Full Weights (Ours) & \textbf{41.5} & \textbf{27.5} & \textbf{48.3} \\ \hline
        \end{tabular}
        }
        \caption{Ablation study on the pre-trained transformer decoder.}
        \label{tab:no_trans_decoder}
    \end{minipage}
    \hspace{1.5mm}
    \begin{minipage}[t]{0.5\textwidth}
        \centering
        \resizebox{1.\linewidth}{!}{
        \begin{tabular}{l | ccc}
            Backbone & \# of params (M) & VRAM (G) & GPU hours (h) \\
            \midrule
            
            R50 & 47.0 & 33.5 & 100.0 \\
            Swin-T & 50.4 & 41.5 & 100.4 \\ \hline
        \end{tabular}
        }
        \caption{\footnotesize Computational costs of Mask-JEPA.}
    \label{tab:compute_cost}
    \end{minipage}
    \vspace{-25pt}
\end{table}

\nbf{Computational Effectiveness}
We detail the overall computational cost of Mask-JEPA in Table~\ref{tab:compute_cost}. During the training Mask-JEPA, we observed only a slight increase in the number of learnable parameters compared to Mask2Former's 44M. Notably, the target segmentation module is not counted among the learnable parameters as it is updated via EMA. In comparison to Mask2Former, which necessitates 32GB of VRAM, Mask-JEPA exhibits minimal increase in VRAM usage. Given that Mask-JEPA operates with a batch size of 32, versus Mask2Former's 16, this suggests VRAM consumption might actually be more efficient for the same batch size. Furthermore, this method allows for the amortization of costs across various downstream datasets and universal image segmentation tasks within a relatively brief period of 100 GPU hours, markedly shorter than the typically longer durations (exceeding $10^3$ GPU hours~\cite{assran2023self, he2022masked}) common in the field of self-supervised learning.

\nbf{Limitations}
We noticed that the improvement on the MS COCO dataset was not as marked as on ADE20K or Cityscapes. We attribute this to the IN1K dataset (1.3M), used for self-supervised learning, being insufficient size for the MS COCO dataset (328K), in contrast to ADE20K (28K) and Cityscapes (5K). Pretraining with full IN21K (14M) might lead to a significant performance boost.

Furthermore, Mask-JEPA currently does not include a specialized component for training queries.  In the context of MCA, a query can act as an embedding representing each object. Looking ahead, we aim to generate proper pseudo-labels to create initial embedding queries, which will lead to better representations.


\section{Related Works}
\label{sec:Related}

\textbf{Mask Classification Architectures for Universal Image Segmentation.}
MaskFormer~\cite{cheng2021per} introduced a versatile model designed for universal image segmentation tasks, eliminating the need for task-specific models. This innovation was further enhanced by Mask2Former's masked attention block mechanism~\cite{cheng2022masked} and MP-Former's novel approach to mask prediction~\cite{zhang2023mp}. OneFormer~\cite{jain2023oneformer} championed a one-shot training paradigm for true universality, while DFormer~\cite{wang2023dformer} integrated principles from diffusion models. Notably, our Mask-JEPA is compatible with all these methods.

\nbf{Self-Supervised Pretraining for Image Segmentation.}
The lack of annotated data for image segmentation has prompted a shift towards self-supervised pretraining. Yang  \etal.~\cite{yang2022fully} implemented a jigsaw methodology, transforming it into a patch classification paradigm, which resulted in significant improvements. Concurrently, DDeP~\cite{brempong2022decoder} highlighted the challenges of arbitrary decoder initialization, promoting a denoising pretraining strategy. It is important to note that some methods were limited to CNNs or exclusively to semantic segmentation.

\nbf{Unsupervised Image Segmentation.}
The emergence of self-supervised and unsupervised techniques has revitalized various segmentation approaches. Many spotlighted pixel-level comprehension via cross-view consistency~\cite{ji2019invariant,cho2021picie,zhang2021looking,ke2022unsupervised,wen2022self,wang2022fully} and intuitive visual priors~\cite{hwang2019segsort,zhang2020self,van2021unsupervised, seong2023leveraging}. 
Zadaianchuk \etal \cite{zadaianchuk2022unsupervised} leveraged pre-trained object representations, whilst others were inspired by pre-trained generative models~\cite{melas2021finding}. Efforts have explored ViTs coupled with DINO~\cite{caron2021emerging}, emphasizing pixel detail~\cite{hamilton2021unsupervised, yin2022transfgu, melas2022deep, van2022discovering, li2023acseg}. Recently, U2Seg~\cite{niu2024unsupervised} successfully generalized an unsupervised approach to universal image segmentation, while other works remain focused on semantic segmentation. Although some studies report that transfer learning can improve image segmentation performance, it is unclear whether this approach can be generalized to MCAs.
We provide an in-depth explanation of previous studies in Appendix~\ref{supp:extended_related}.

\vspace{-3pt}
\section{Conclusion}
\label{sec:conclusion}

Driven by recent MCA, we introduce Mask-JEPA, a joint embedding predictive architecture tailored for mask classfication architecture. Our findings reveal that Mask-JEPA adeptly captures both semantic objects and edges, demonstrating the capabilities of MCA. This inherent trait empowers Mask-JEPA to enhance the performance of universal image segmentation across three large-scale datasets (ADE20K, Cityscapes, and MS COCO) under various scenarios. 
We believe that our method has potentials to innovative self-supervised training especially for universal image segmentation models.


%
%
\bibliographystyle{splncs04}
\bibliography{_main}

\begin{thebibliography}{10}
\providecommand{\url}[1]{\texttt{#1}}
\providecommand{\urlprefix}{URL }
\providecommand{\doi}[1]{https://doi.org/#1}

\bibitem{assran2023self}
Assran, M., Duval, Q., Misra, I., Bojanowski, P., Vincent, P., Rabbat, M., LeCun, Y., Ballas, N.: Self-supervised learning from images with a joint-embedding predictive architecture. In: Proceedings of the IEEE/CVF Conference on Computer Vision and Pattern Recognition. pp. 15619--15629 (2023)

\bibitem{ba2016layer}
Ba, J.L., Kiros, J.R., Hinton, G.E.: Layer normalization. arXiv preprint arXiv:1607.06450  (2016)

\bibitem{bao2021beit}
Bao, H., Dong, L., Piao, S., Wei, F.: Beit: Bert pre-training of image transformers. In: International Conference on Learning Representations (2021)

\bibitem{bardes2023v}
Bardes, A., Garrido, Q., Ponce, J., Chen, X., Rabbat, M., LeCun, Y., Assran, M., Ballas, N.: V-jepa: Latent video prediction for visual representation learning  (2023)

\bibitem{bardes2022vicregl}
Bardes, A., Ponce, J., LeCun, Y.: Vicregl: Self-supervised learning of local visual features. Advances in Neural Information Processing Systems  \textbf{35},  8799--8810 (2022)

\bibitem{bardes2023mc}
Bardes, A., Ponce, J., LeCun, Y.: Mc-jepa: A joint-embedding predictive architecture for self-supervised learning of motion and content features. arXiv preprint arXiv:2307.12698  (2023)

\bibitem{bielski2022move}
Bielski, A., Favaro, P.: Move: Unsupervised movable object segmentation and detection. Advances in Neural Information Processing Systems  \textbf{35},  33371--33386 (2022)

\bibitem{bordes2022high}
Bordes, F., Balestriero, R., Vincent, P.: High fidelity visualization of what your self-supervised representation knows about. Transactions on Machine Learning Research  (2022)

\bibitem{brempong2022decoder}
Brempong, E.A., Kornblith, S., Chen, T., Parmar, N., Minderer, M., Norouzi, M.: Decoder denoising pretraining for semantic segmentation. Transactions on Machine Learning Research  (2022)

\bibitem{carion2020end}
Carion, N., Massa, F., Synnaeve, G., Usunier, N., Kirillov, A., Zagoruyko, S.: End-to-end object detection with transformers. In: European conference on computer vision. pp. 213--229. Springer (2020)

\bibitem{caron2020unsupervised}
Caron, M., Misra, I., Mairal, J., Goyal, P., Bojanowski, P., Joulin, A.: Unsupervised learning of visual features by contrasting cluster assignments. Advances in neural information processing systems  \textbf{33},  9912--9924 (2020)

\bibitem{caron2021emerging}
Caron, M., Touvron, H., Misra, I., J{\'e}gou, H., Mairal, J., Bojanowski, P., Joulin, A.: Emerging properties in self-supervised vision transformers. In: Proceedings of the IEEE/CVF international conference on computer vision. pp. 9650--9660 (2021)

\bibitem{chen2021transunet}
Chen, J., Lu, Y., Yu, Q., Luo, X., Adeli, E., Wang, Y., Lu, L., Yuille, A.L., Zhou, Y.: Transunet: Transformers make strong encoders for medical image segmentation. arXiv preprint arXiv:2102.04306  (2021)

\bibitem{chen2016semantic}
Chen, L.C., Barron, J.T., Papandreou, G., Murphy, K., Yuille, A.L.: Semantic image segmentation with task-specific edge detection using cnns and a discriminatively trained domain transform. In: Proceedings of the IEEE conference on computer vision and pattern recognition. pp. 4545--4554 (2016)

\bibitem{chen2018encoder}
Chen, L.C., Zhu, Y., Papandreou, G., Schroff, F., Adam, H.: Encoder-decoder with atrous separable convolution for semantic image segmentation. In: Proceedings of the European conference on computer vision (ECCV). pp. 801--818 (2018)

\bibitem{chen2020simple}
Chen, T., Kornblith, S., Norouzi, M., Hinton, G.: A simple framework for contrastive learning of visual representations. In: International conference on machine learning. pp. 1597--1607. PMLR (2020)

\bibitem{cheng2020panoptic}
Cheng, B., Collins, M.D., Zhu, Y., Liu, T., Huang, T.S., Adam, H., Chen, L.C.: Panoptic-deeplab: A simple, strong, and fast baseline for bottom-up panoptic segmentation. In: Proceedings of the IEEE/CVF conference on computer vision and pattern recognition. pp. 12475--12485 (2020)

\bibitem{cheng2022masked}
Cheng, B., Misra, I., Schwing, A.G., Kirillov, A., Girdhar, R.: Masked-attention mask transformer for universal image segmentation. In: Proceedings of the IEEE/CVF conference on computer vision and pattern recognition. pp. 1290--1299 (2022)

\bibitem{cheng2021per}
Cheng, B., Schwing, A., Kirillov, A.: Per-pixel classification is not all you need for semantic segmentation. Advances in Neural Information Processing Systems  \textbf{34},  17864--17875 (2021)

\bibitem{cho2021picie}
Cho, J.H., Mall, U., Bala, K., Hariharan, B.: Picie: Unsupervised semantic segmentation using invariance and equivariance in clustering. In: Proceedings of the IEEE/CVF Conference on Computer Vision and Pattern Recognition. pp. 16794--16804 (2021)

\bibitem{cordts2016cityscapes}
Cordts, M., Omran, M., Ramos, S., Rehfeld, T., Enzweiler, M., Benenson, R., Franke, U., Roth, S., Schiele, B.: The cityscapes dataset for semantic urban scene understanding. In: Proceedings of the IEEE conference on computer vision and pattern recognition. pp. 3213--3223 (2016)

\bibitem{cordts2015cityscapes}
Cordts, M., Omran, M., Ramos, S., Scharw{\"a}chter, T., Enzweiler, M., Benenson, R., Franke, U., Roth, S., Schiele, B.: The cityscapes dataset. In: CVPR Workshop on the Future of Datasets in Vision. vol.~2. sn (2015)

\bibitem{dosovitskiy2020image}
Dosovitskiy, A., Beyer, L., Kolesnikov, A., Weissenborn, D., Zhai, X., Unterthiner, T., Dehghani, M., Minderer, M., Heigold, G., Gelly, S., et~al.: An image is worth 16x16 words: Transformers for image recognition at scale. In: International Conference on Learning Representations (2020)

\bibitem{garrido2024learning}
Garrido, Q., Assran, M., Ballas, N., Bardes, A., Najman, L., LeCun, Y.: Learning and leveraging world models in visual representation learning. arXiv preprint arXiv:2403.00504  (2024)

\bibitem{grill2020bootstrap}
Grill, J.B., Strub, F., Altch{\'e}, F., Tallec, C., Richemond, P., Buchatskaya, E., Doersch, C., Avila~Pires, B., Guo, Z., Gheshlaghi~Azar, M., et~al.: Bootstrap your own latent-a new approach to self-supervised learning. Advances in neural information processing systems  \textbf{33},  21271--21284 (2020)

\bibitem{hamilton2021unsupervised}
Hamilton, M., Zhang, Z., Hariharan, B., Snavely, N., Freeman, W.T.: Unsupervised semantic segmentation by distilling feature correspondences. In: International Conference on Learning Representations (2021)

\bibitem{he2022masked}
He, K., Chen, X., Xie, S., Li, Y., Doll{\'a}r, P., Girshick, R.: Masked autoencoders are scalable vision learners. In: Proceedings of the IEEE/CVF conference on computer vision and pattern recognition. pp. 16000--16009 (2022)

\bibitem{he2020momentum}
He, K., Fan, H., Wu, Y., Xie, S., Girshick, R.: Momentum contrast for unsupervised visual representation learning. In: Proceedings of the IEEE/CVF conference on computer vision and pattern recognition. pp. 9729--9738 (2020)

\bibitem{he2016deep}
He, K., Zhang, X., Ren, S., Sun, J.: Deep residual learning for image recognition. In: Proceedings of the IEEE conference on computer vision and pattern recognition. pp. 770--778 (2016)

\bibitem{ho2020denoising}
Ho, J., Jain, A., Abbeel, P.: Denoising diffusion probabilistic models. Advances in neural information processing systems  \textbf{33},  6840--6851 (2020)

\bibitem{huang2022segdiscover}
Huang, H., Chen, Z., Rudin, C.: Segdiscover: Visual concept discovery via unsupervised semantic segmentation. arXiv preprint arXiv:2204.10926  (2022)

\bibitem{hwang2019segsort}
Hwang, J.J., Yu, S.X., Shi, J., Collins, M.D., Yang, T.J., Zhang, X., Chen, L.C.: Segsort: Segmentation by discriminative sorting of segments. In: Proceedings of the IEEE/CVF International Conference on Computer Vision. pp. 7334--7344 (2019)

\bibitem{jain2023oneformer}
Jain, J., Li, J., Chiu, M.T., Hassani, A., Orlov, N., Shi, H.: Oneformer: One transformer to rule universal image segmentation. In: Proceedings of the IEEE/CVF Conference on Computer Vision and Pattern Recognition. pp. 2989--2998 (2023)

\bibitem{ji2019invariant}
Ji, X., Henriques, J.F., Vedaldi, A.: Invariant information clustering for unsupervised image classification and segmentation. In: Proceedings of the IEEE/CVF international conference on computer vision. pp. 9865--9874 (2019)

\bibitem{ke2022unsupervised}
Ke, T.W., Hwang, J.J., Guo, Y., Wang, X., Yu, S.X.: Unsupervised hierarchical semantic segmentation with multiview cosegmentation and clustering transformers. In: Proceedings of the IEEE/CVF Conference on Computer Vision and Pattern Recognition. pp. 2571--2581 (2022)

\bibitem{kirillov2019panoptic}
Kirillov, A., He, K., Girshick, R., Rother, C., Doll{\'a}r, P.: Panoptic segmentation. In: Proceedings of the IEEE/CVF conference on computer vision and pattern recognition. pp. 9404--9413 (2019)

\bibitem{krahenbuhl2011efficient}
Kr{\"a}henb{\"u}hl, P., Koltun, V.: Efficient inference in fully connected crfs with gaussian edge potentials. Advances in neural information processing systems  \textbf{24} (2011)

\bibitem{lecun2022path}
LeCun, Y.: A path towards autonomous machine intelligence version 0.9. 2, 2022-06-27. Open Review  \textbf{62} (2022)

\bibitem{li2023mask}
Li, F., Zhang, H., Xu, H., Liu, S., Zhang, L., Ni, L.M., Shum, H.Y.: Mask dino: Towards a unified transformer-based framework for object detection and segmentation. In: Proceedings of the IEEE/CVF Conference on Computer Vision and Pattern Recognition. pp. 3041--3050 (2023)

\bibitem{li2023acseg}
Li, K., Wang, Z., Cheng, Z., Yu, R., Zhao, Y., Song, G., Liu, C., Yuan, L., Chen, J.: Acseg: Adaptive conceptualization for unsupervised semantic segmentation. In: Proceedings of the IEEE/CVF Conference on Computer Vision and Pattern Recognition. pp. 7162--7172 (2023)

\bibitem{li2024t}
Li, L., Xue, H., Song, Y., Salim, F.: T-jepa: A joint-embedding predictive architecture for trajectory similarity computation. arXiv preprint arXiv:2406.12913  (2024)

\bibitem{li2020improving}
Li, X., Li, X., Zhang, L., Cheng, G., Shi, J., Lin, Z., Tan, S., Tong, Y.: Improving semantic segmentation via decoupled body and edge supervision. In: Computer Vision--ECCV 2020: 16th European Conference, Glasgow, UK, August 23--28, 2020, Proceedings, Part XVII 16. pp. 435--452. Springer (2020)

\bibitem{lin2017feature}
Lin, T.Y., Doll{\'a}r, P., Girshick, R., He, K., Hariharan, B., Belongie, S.: Feature pyramid networks for object detection. In: Proceedings of the IEEE conference on computer vision and pattern recognition. pp. 2117--2125 (2017)

\bibitem{lin2014microsoft}
Lin, T.Y., Maire, M., Belongie, S., Hays, J., Perona, P., Ramanan, D., Doll{\'a}r, P., Zitnick, C.L.: Microsoft coco: Common objects in context. In: Computer Vision--ECCV 2014: 13th European Conference, Zurich, Switzerland, September 6-12, 2014, Proceedings, Part V 13. pp. 740--755. Springer (2014)

\bibitem{liu2021swin}
Liu, Z., Lin, Y., Cao, Y., Hu, H., Wei, Y., Zhang, Z., Lin, S., Guo, B.: Swin transformer: Hierarchical vision transformer using shifted windows. In: Proceedings of the IEEE/CVF international conference on computer vision. pp. 10012--10022 (2021)

\bibitem{loshchilov2018decoupled}
Loshchilov, I., Hutter, F.: Decoupled weight decay regularization. In: International Conference on Learning Representations (2018)

\bibitem{melas2021finding}
Melas-Kyriazi, L., Rupprecht, C., Laina, I., Vedaldi, A.: Finding an unsupervised image segmenter in each of your deep generative models. In: International Conference on Learning Representations (2021)

\bibitem{melas2022deep}
Melas-Kyriazi, L., Rupprecht, C., Laina, I., Vedaldi, A.: Deep spectral methods: A surprisingly strong baseline for unsupervised semantic segmentation and localization. In: Proceedings of the IEEE/CVF Conference on Computer Vision and Pattern Recognition. pp. 8364--8375 (2022)

\bibitem{niu2024unsupervised}
Niu, D., Wang, X., Han, X., Lian, L., Herzig, R., Darrell, T.: Unsupervised universal image segmentation. In: Proceedings of the IEEE/CVF Conference on Computer Vision and Pattern Recognition. pp. 22744--22754 (2024)

\bibitem{park2022self}
Park, N., Kim, W., Heo, B., Kim, T., Yun, S.: What do self-supervised vision transformers learn? In: The Eleventh International Conference on Learning Representations (2022)

\bibitem{russakovsky2015imagenet}
Russakovsky, O., Deng, J., Su, H., Krause, J., Satheesh, S., Ma, S., Huang, Z., Karpathy, A., Khosla, A., Bernstein, M., et~al.: Imagenet large scale visual recognition challenge. International journal of computer vision  \textbf{115},  211--252 (2015)

\bibitem{seong2023leveraging}
Seong, H.S., Moon, W., Lee, S., Heo, J.P.: Leveraging hidden positives for unsupervised semantic segmentation. In: Proceedings of the IEEE/CVF Conference on Computer Vision and Pattern Recognition. pp. 19540--19549 (2023)

\bibitem{song2020denoising}
Song, J., Meng, C., Ermon, S.: Denoising diffusion implicit models. In: International Conference on Learning Representations (2020)

\bibitem{van2021unsupervised}
Van~Gansbeke, W., Vandenhende, S., Georgoulis, S., Van~Gool, L.: Unsupervised semantic segmentation by contrasting object mask proposals. In: Proceedings of the IEEE/CVF International Conference on Computer Vision. pp. 10052--10062 (2021)

\bibitem{van2022discovering}
Van~Gansbeke, W., Vandenhende, S., Van~Gool, L.: Discovering object masks with transformers for unsupervised semantic segmentation. arXiv preprint arXiv:2206.06363  (2022)

\bibitem{vaswani2017attention}
Vaswani, A., Shazeer, N., Parmar, N., Uszkoreit, J., Jones, L., Gomez, A.N., Kaiser, {\L}., Polosukhin, I.: Attention is all you need. Advances in neural information processing systems  \textbf{30} (2017)

\bibitem{vincent2011connection}
Vincent, P.: A connection between score matching and denoising autoencoders. Neural computation  \textbf{23}(7),  1661--1674 (2011)

\bibitem{vincent2008extracting}
Vincent, P., Larochelle, H., Bengio, Y., Manzagol, P.A.: Extracting and composing robust features with denoising autoencoders. In: Proceedings of the 25th international conference on Machine learning. pp. 1096--1103 (2008)

\bibitem{wang2023dformer}
Wang, H., Cao, J., Anwer, R.M., Xie, J., Khan, F.S., Pang, Y.: Dformer: Diffusion-guided transformer for universal image segmentation. arXiv preprint arXiv:2306.03437  (2023)

\bibitem{wang2022freesolo}
Wang, X., Yu, Z., De~Mello, S., Kautz, J., Anandkumar, A., Shen, C., Alvarez, J.M.: Freesolo: Learning to segment objects without annotations. In: Proceedings of the IEEE/CVF Conference on Computer Vision and Pattern Recognition. pp. 14176--14186 (2022)

\bibitem{wang2022fully}
Wang, Y., Zhuo, W., Li, Y., Wang, Z., Ju, Q., Zhu, W.: Fully self-supervised learning for semantic segmentation. arXiv preprint arXiv:2202.11981  (2022)

\bibitem{wen2022self}
Wen, X., Zhao, B., Zheng, A., Zhang, X., Qi, X.: Self-supervised visual representation learning with semantic grouping. Advances in Neural Information Processing Systems  \textbf{35},  16423--16438 (2022)

\bibitem{wu2022denoising}
Wu, Q., Ye, H., Gu, Y., Zhang, H., Wang, L., He, D.: Denoising masked autoencoders help robust classification. In: The Eleventh International Conference on Learning Representations (2022)

\bibitem{xia2017w}
Xia, X., Kulis, B.: W-net: A deep model for fully unsupervised image segmentation. arXiv preprint arXiv:1711.08506  (2017)

\bibitem{xie2021self}
Xie, Z., Lin, Y., Yao, Z., Zhang, Z., Dai, Q., Cao, Y., Hu, H.: Self-supervised learning with swin transformers. arXiv preprint arXiv:2105.04553  (2021)

\bibitem{xie2022simmim}
Xie, Z., Zhang, Z., Cao, Y., Lin, Y., Bao, J., Yao, Z., Dai, Q., Hu, H.: Simmim: A simple framework for masked image modeling. In: Proceedings of the IEEE/CVF Conference on Computer Vision and Pattern Recognition. pp. 9653--9663 (2022)

\bibitem{yang2022fully}
Yang, Z., Yu, H., He, Y., Sun, W., Mao, Z.H., Mian, A.: Fully convolutional network-based self-supervised learning for semantic segmentation. IEEE Transactions on Neural Networks and Learning Systems  (2022)

\bibitem{yin2022transfgu}
Yin, Z., Wang, P., Wang, F., Xu, X., Zhang, H., Li, H., Jin, R.: Transfgu: a top-down approach to fine-grained unsupervised semantic segmentation. In: European conference on computer vision. pp. 73--89. Springer (2022)

\bibitem{yuan2020segfix}
Yuan, Y., Xie, J., Chen, X., Wang, J.: Segfix: Model-agnostic boundary refinement for segmentation. In: Computer Vision--ECCV 2020: 16th European Conference, Glasgow, UK, August 23--28, 2020, Proceedings, Part XII 16. pp. 489--506. Springer (2020)

\bibitem{zadaianchuk2022unsupervised}
Zadaianchuk, A., Kleindessner, M., Zhu, Y., Locatello, F., Brox, T.: Unsupervised semantic segmentation with self-supervised object-centric representations. In: The Eleventh International Conference on Learning Representations (2022)

\bibitem{zhang2021looking}
Zhang, F., Torr, P., Ranftl, R., Richter, S.: Looking beyond single images for contrastive semantic segmentation learning. Advances in neural information processing systems  \textbf{34},  3285--3297 (2021)

\bibitem{zhang2023mp}
Zhang, H., Li, F., Xu, H., Huang, S., Liu, S., Ni, L.M., Zhang, L.: Mp-former: Mask-piloted transformer for image segmentation. In: Proceedings of the IEEE/CVF Conference on Computer Vision and Pattern Recognition. pp. 18074--18083 (2023)

\bibitem{zhang2020self}
Zhang, X., Maire, M.: Self-supervised visual representation learning from hierarchical grouping. Advances in Neural Information Processing Systems  \textbf{33},  16579--16590 (2020)

\bibitem{zhao2017pyramid}
Zhao, H., Shi, J., Qi, X., Wang, X., Jia, J.: Pyramid scene parsing network. In: Proceedings of the IEEE conference on computer vision and pattern recognition. pp. 2881--2890 (2017)

\bibitem{zhong2021pixel}
Zhong, Y., Yuan, B., Wu, H., Yuan, Z., Peng, J., Wang, Y.X.: Pixel contrastive-consistent semi-supervised semantic segmentation. In: Proceedings of the IEEE/CVF International Conference on Computer Vision. pp. 7273--7282 (2021)

\bibitem{zhou2019semantic}
Zhou, B., Zhao, H., Puig, X., Xiao, T., Fidler, S., Barriuso, A., Torralba, A.: Semantic understanding of scenes through the ade20k dataset. International Journal of Computer Vision  \textbf{127},  302--321 (2019)

\bibitem{zhu2020deformable}
Zhu, X., Su, W., Lu, L., Li, B., Wang, X., Dai, J.: Deformable detr: Deformable transformers for end-to-end object detection. In: International Conference on Learning Representations (2020)

\bibitem{ziegler2022self}
Ziegler, A., Asano, Y.M.: Self-supervised learning of object parts for semantic segmentation. In: Proceedings of the IEEE/CVF Conference on Computer Vision and Pattern Recognition. pp. 14502--14511 (2022)

\end{thebibliography}

\renewcommand{\thesection}{\Alph{section}}
\renewcommand{\thefigure}{\Alph{figure}}
\renewcommand{\thetable}{\Alph{table}}
\setcounter{section}{0}

\section{Implementation Details}
\label{supp:imp_detail}

For the Mask-JEPA, we utilized 8 NVIDIA RTX 4090 GPUs, each with 24GB of memory. The learning rate was set to 0.0001, using the AdamW~\cite{loshchilov2018decoupled} optimizer, and we increased the batch size to 32. For training the MP-Former~\cite{zhang2023mp} and OneFormer~\cite{jain2023oneformer} models, 4 NVIDIA V100 GPUs were used, each having a 32GB memory. The same training procedure was applied to fine-tune the Mask2Former, MP-Former, and OneFormer models, with the initial learning rate adjusted to 0.0002. All other hyperparameters remained consistent with the original implementations of each model.

Regarding the Exponential Moving Average (EMA), we adopted the following equation:
\begin{equation}
\bar{\theta} \leftarrow \tau \bar{\theta} + (1 - \tau) \theta
\end{equation}
Here, $\tau$ is incrementally increased from 0.996 to 1 linearly over the training steps.

\section{Detailed Related Works}
\label{supp:extended_related}
\nbf{Mask Classification Architectures for Universal Image Segmentation.} Recent advancements in universal image segmentation have primarily focused on formulating unified frameworks to efficiently address tasks like semantic, instance, and panoptic segmentation. The MaskFormer~\cite{cheng2021per} approach, for instance, utilized mask classification for both semantic and instance-level tasks, outperforming traditional per-pixel classification methods. Mask2Former~\cite{cheng2022masked} further enhanced this approach by introducing masked attention for localized feature extraction, while its successor, the mask-piloted Transformer~\cite{zhang2023mp}, optimized mask predictions between consecutive decoder layers.

However, the OneFormer~\cite{jain2023oneformer} model revolutionized the field with its train-once design. It integrated a task-conditioned joint training strategy and dynamic task determination, ensuring superior performance across various segmentation tasks without the need for individual, task-specific training. Meanwhile, the DFormer's approach~\cite{wang2023dformer}, which treats segmentation as a denoising process using a diffusion model, has pushed the boundaries of universal segmentation and set new benchmarks. Our Mask-JEPA seamlessly incorporates the principles of these mask classification architectures.

\nbf{Self-Supervised Pretraining for Image Segmentation.}
Several key studies have contributed significantly to the methodology of self-supervised pretraining for image segmentation. DDeP~\cite{brempong2022decoder} diverges from traditional practices that often involve randomly initializing the decoder in segmentation models. This study introduces a denoising pretraining for the decoder, which complements the supervised pretraining of the encoder, thereby enhancing the overall effectiveness of segmentation models.

In the research conducted by Yang \emph{et al.}~\cite{yang2022fully}, a novel framework is presented, conceptualizing the self-supervised learning process as a jigsaw puzzle problem, addressed using a fully convolutional network. This methodology makes effective use of unlabeled data in training semantic segmentation models. Moreover, they propose a bootstrapped training scheme, incorporating a pyramid-global-guided strategy and a context-aware embedding module to utilize global semantic knowledge for self-supervision. However, these methods are primarily applicable to CNN architectures and focus predominantly on semantic segmentation tasks.

\nbf{Joint Embedding Predictive Architectures.} Since LeCun introduced the initial idea of Joint Embedding Predictive Architectures (JEPA)~\cite{lecun2022path}, JEPA has shown promising results in self-supervised learning across various domains. I-JEPA~\cite{assran2023self} for images and V-JEPA~\cite{bardes2023v} for videos have both achieved state-of-the-art performance on downstream tasks. T-JEPA~\cite{li2024t} was proposed for trajectory similarity computation, and MC-JEPA~\cite{bardes2023mc} jointly learns optical flow and content features. Garrido \etal~\cite{garrido2024learning} extended JEPA to predict global photometric transformations with Image World Models. These works demonstrate JEPA's effectiveness in learning rich, task-agnostic representations across different domains, and our Mask-JEPA successfully applies the JEPA concept to MCA.

\nbf{Unsupervised Semantic Segmentation.}
Unsupervised semantic segmentation has garnered significant attention, evolving with the development of self-supervised and unsupervised learning methodologies. Traditional approaches such as early CRF models focused on maximizing label agreement between similar pixels based on low-level appearance information, using simple adjacency definitions like 4-connected or 8-connected grids \cite{krahenbuhl2011efficient, xia2017w}. However, these methods often fell short in capturing high-level semantic information in images. In contrast, recent advancements have emphasized pixel-level self-supervised representation learning, utilizing cross-view consistency \cite{ji2019invariant,cho2021picie,zhang2021looking,ziegler2022self,ke2022unsupervised,wen2022self,wang2022fully}, visual priors \cite{hwang2019segsort,zhang2020self,van2021unsupervised}, and continuity of video frames \cite{bielski2022move}. Notable approaches include methods leveraging pre-trained object-centric representations and generative models \cite{zadaianchuk2022unsupervised, melas2021finding}, as well as the utilization of self-supervised pre-trained CNNs \cite{wang2022freesolo,huang2022segdiscover}.

The use of self-supervised Vision Transformers (ViTs)~\cite{dosovitskiy2020image}, particularly DINO~\cite{caron2021emerging}, has been explored for unsupervised dense prediction tasks due to their ability to represent pixel-level semantic relationships. Techniques like STEGO \cite{hamilton2021unsupervised} have trained segmentation heads by distilling feature correspondences, and TransFGU \cite{yin2022transfgu} has extracted class activate maps from DINO models. These methods aim at forming compact clusters of pixel features and learning better pixel-level representations. Similarly, spectral decomposition on affinity graphs \cite{melas2022deep} and methods like MaskDistill \cite{van2022discovering} have been utilized for segmenting images into regions based on pixel-level representations and mask priors, primarily focusing on foreground object segmentation. ACSeg~\cite{li2023acseg} adaptively map learnable prototypes to image-specific concepts, optimized with a modularity loss for scene complexity.

Unsupervised semantic segmentation methods aim to group semantically meaningful pixels without any labels. Works such as those \cite{hwang2019segsort, zhong2021pixel, zhang2021looking, wen2022self} have demonstrated that fine-tuning with labeled segmentation datasets can improve mIoU scores. However, it remains unclear whether these approaches are effective for mask classification architectures. Furthermore, they lack specific components necessary for the proper training of a transformer decoder. In contrast, Mask-JEPA is capable of training full mask classification architectures, including the pixel decoder.

\section{Extended Results}
\label{supp:extended_results}

\subsection{Denoising Target}
\label{supp:denoising_target}

\begin{table}[t!]\setlength{\tabcolsep}{10pt}
  \centering
  \resizebox{0.35\linewidth}{!}{
    \begin{tabular}{l | c}

        Target & mIoU \\
        \midrule
        
        Raw Image & 48.0 \\

        Gaussian Noise & \textbf{48.3} \\
    \end{tabular}
    }
  \caption{\textbf{Denoising Target.} Raw image vs. Gaussian Noise.
      }
    \label{tab:target}
\end{table}

We investigate the importance of denoising the prediction target (raw images vs. Gaussian Noise) in Table~\ref{tab:target}. The results indicate that predicting Gaussian noise can lead to better performance, achieving a +0.3 improvement in mIoU.

\subsection{No ImageNet Pretrained Backbone Weights Regimes}
\label{supp:no_ignt}

\begin{table}[t!]\setlength{\tabcolsep}{10pt}
  \centering
  \resizebox{0.55\linewidth}{!}{
    \begin{tabular}{l | ccc}

        Method & PQ & AP & mIoU \\
        \midrule
        
        Mask2Foremr & 30.3 & 19.5 & 40.3 \\

        \quad \textbf{+ Mask-JEPA} & \textbf{32.7} & \textbf{21.3} & \textbf{40.7} \\
    \end{tabular}
    }
  \caption{Image segmentation results on ADE20K \texttt{val} without IN1K pretrained backbone weights.}
    \label{tab:without_imgt}
\end{table}

We investigate the performance of Mask-JEPA in the context where classification-supervised pretrained backbone weights are inaccessible.

\noindent
\textbf{Pretraining:}
We pretrained Mask-JEPA using Mask2Former with ImageNet (IN1K) for 5 epochs. 
We used the random initialized whole backbone, pixel decoder, transformer decoder weights.

\noindent
\textbf{Setup:} We compared the performance gain between Mask-JEPA pretrained weights and randomly initialized weights on the ADE20K datasets for 160k iterations.

\noindent
\textbf{Results:} Table~\ref{tab:without_imgt} shows the results, demonstrating improvements across all metrics used to evaluate universal image segmentation. These findings indicate that Mask-JEPA is not only effective in scenarios where access to an ImageNet classification pretrained backbone is unavailable, but it also has the potential to adapt to distinctly different image distributions, such as the medical domain.

\subsection{Pretrained with COCO Dataset}
\label{supp:coco_pretrain}

\begin{table}[t!]\setlength{\tabcolsep}{10pt}
  \centering
  \resizebox{0.44\linewidth}{!}{
    \begin{tabular}{l | ccc}

        Dataset & PQ & AP & mIoU \\
        \midrule
        
        IN1K & \textbf{41.5} & 27.5 & \textbf{48.3} \\

        COCO & 41.4 & \textbf{27.8} & 47.0 \\
    \end{tabular}
    }
  \caption{Image segmentation results on ADE20K \texttt{val} with fine-tuning. Weights pre-trained using Mask-JEPA with IN1K and COCO datasets.}
    \label{tab:coco_pretrain}
\end{table}
We investigate the performance of Mask-JEPA in the context where ImageNet dataset is inaccessible and when we have to utilize another dataset like MS-COCO.\\
\nbf{Pretraining:}
Mask-JEPA was pretrained using Mask2Former on the COCO \texttt{train2017} dataset for 187,500 iterations, matching the duration used for IN1K over 5 epochs. 
We used the pretrained backbone weights from ImageNet classification and kept them frozen.\\
\nbf{Setup:}
We assessed the performance by comparing Mask-JEPA pretrained on the IN1K and COCO datasets, with fine-tuning conducted on the ADE20K dataset for 160k iterations. \\
\nbf{Result:}
Table~\ref{tab:coco_pretrain} indicates that pretraining with the COCO dataset is competitive with that of IN1K pretrained weights. However, for the semantic segmentation task, there wasn't a significant improvement observed. We interpret this phenomenon as an indication that Mask-JEPA may struggle to generalize across the same class but different instances with relatively fewer images.

\subsection{Increase Pretraining Epochs}
\label{supp:increase_epochs}

\begin{table}[t!]\setlength{\tabcolsep}{10pt}
  \centering
  \resizebox{0.44\linewidth}{!}{
    \begin{tabular}{l | ccc}

        Epochs & PQ & AP & mIoU \\
        \midrule
        
        5 & \textbf{41.5} & 27.5 & \textbf{48.3} \\

        10 & 41.1 & 27.5 & 47.5 \\

        15 & 41.1 & 27.6 & 47.3 \\

        20 & 40.5 & \textbf{28.0} & 47.8 \\
    \end{tabular}
    }
  \caption{Image segmentation results on ADE20K \texttt{val} finetuning with gradually increased pretrained eopchs.}
    \label{tab:increase_epochs}
\end{table}
We investigate the performance of Mask-JEPA in the context where pretraining epochs are increased. \\
\nbf{Pretraining:}
We pretrained Mask-JEPA using Mask2Former with ImageNet (IN1K) for 20 epochs.
We used the pretrained backbone weights from ImageNet classification and kept them frozen. \\
\nbf{Setup:}
We assessed the performance by comparing Mask-JEPA with fine-tuning conducted on the ADE20K dataset for 160k iterations. \\
\nbf{Result:}
Table~\ref{tab:increase_epochs} shows the results of pretraining using Mask-JEPA for up to 20 epochs. This indicates that there is no significant correlation between performance and the number of epochs beyond 5. We believe this is due to the limited capacity of ResNet50 to extract representations. If we use a backbone capable of extracting richer features, we expect the performance to improve as the number of epochs increases.

\subsection{Comparison with Self-Supervised Learning Backbone}
\label{supp:backbone_ssl}

Our objective in this work was to learn useful representations through self-supervised learning using the entire MCA as an unlabeled image. However, the performance of MCA when utilizing weights trained through self-supervised learning for its backbone remains an uncharted territory. We employed the backbone weights pretrained on IN1K using VicRegL~\cite{bardes2022vicregl} and MoBY~\cite{xie2021self} for fine-tuning the ResNet50 and Swin-T backbone Mask2Former respectively, and the results are presented in Table~\ref{tab:backbone_ssl}.

The outcomes indicate that Mask-JEPA possesses considerable potential to achieve full self-supervision without relying on labels from the IN1K classification dataset. It is noteworthy that most of our experiments were conducted on the IN1K classification, employing supervised backbone weights.

\subsection{Extended Qualitative Results}
\label{supp:extended_qual}

\begin{figure*}[!t]
    \centering
    \includegraphics[width=\linewidth]{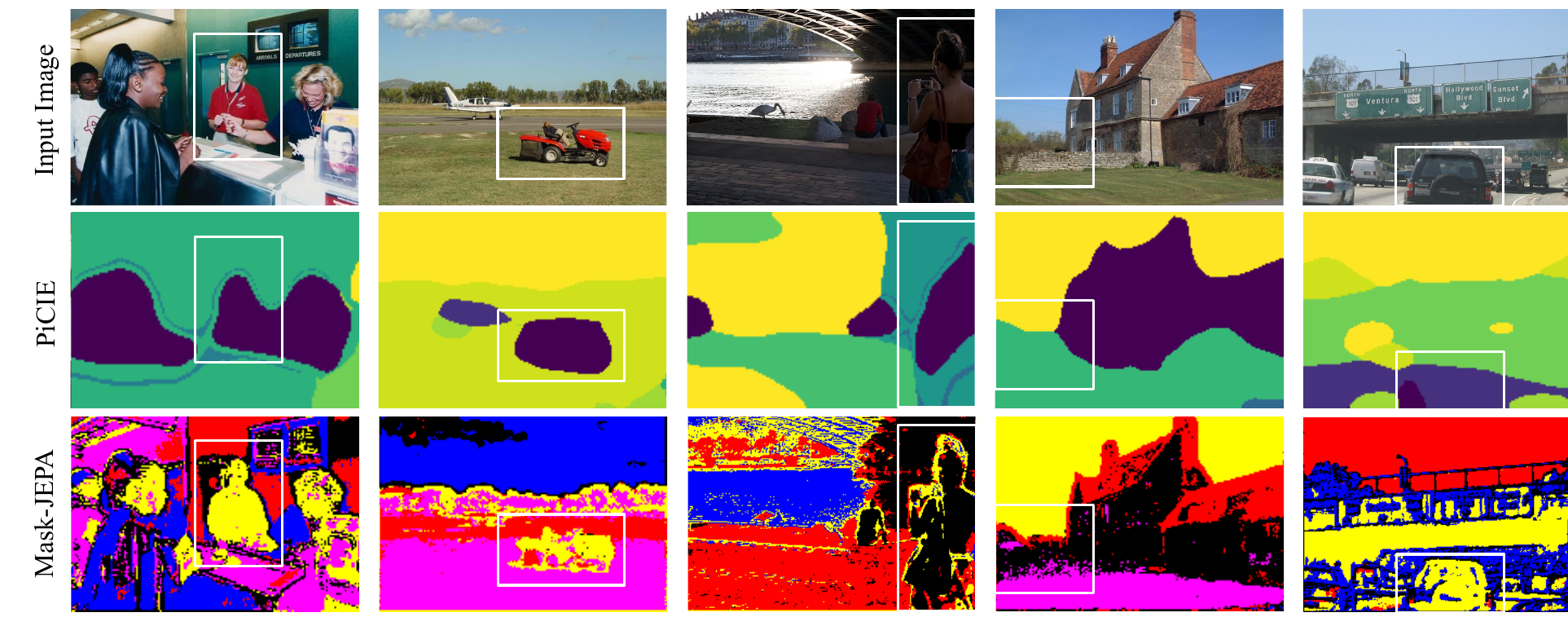}
    \vspace{-20pt}
    \caption{\textbf{Unsupervised Semantic Segmentation Methods vs. Mask-JEPA.} This figure contrasts a unsupervised semantic segmentation method (PiCIE) with our Mask-JEPA, highlighting their performance in object and edge detection.}
    \label{fig:supp_cluster}
\end{figure*}

\begin{table}[t!]\setlength{\tabcolsep}{10pt}
  \centering
  \resizebox{0.95\linewidth}{!}{
    \begin{tabular}{l| l |l | ccc}

        Backbone & Pretrained Backbone & PD+TD Pretrain & PQ & AP & mIoU \\
        \midrule
        
        \multirow{4}{*}{R50}%

        & VICRegL & None & 37.7 & 24.7 & 44.9 \\

        & VICRegL & Mask-JEPA & 38.8 & 26.4 & 45.8 \\

        & IN1K classification & None & 39.7 & 26.4 & 46.6 \\

        & IN1K classification & Mask-JEPA & \textbf{41.5} & \textbf{27.5} & \textbf{48.3} \\

        \midrule
        
        \multirow{4}{*}{Swin-T}%

        & MoBY & None & 41.2& 28.0 & 48.5 \\

        & MoBY & Mask-JEPA & 40.4 & 28.1 & 49.1 \\

        & IN1K classification & None & 40.2 & 27.2 & 47.7 \\

        & IN1K classification & Mask-JEPA & \textbf{41.8} & \textbf{28.5} & \textbf{50.5} \\\hline
    \end{tabular}
    }
  \caption{Segmentation performance on ADE20K validation set using a fine-tuned model with a MoBY-pretrained self-supervised backbone.}
    \label{tab:backbone_ssl}
\end{table}

\subsubsection{Compare to Unsupervised Semantic Segmentation}
Unsupervised semantic segmentation tasks, akin to our work, aim to extract semantic representations. In order to compare these approaches, we visualized methods from unsupervised semantic segmentation~\cite{cho2021picie} alongside our Mask-JEPA in Figure~\ref{fig:supp_cluster}. Our analysis reveals that a unsupervised semantic segmentation method often face challenges in capturing two crucial elements simultaneously: objects and edges. These elements, as inferred from the mask ($\mathcal{F}_{mask}$), are vital in mask classification architectures.

\subsubsection{Additional Visualization}
\begin{figure*}[tp]
    \centering
    \includegraphics[width=0.95\linewidth]{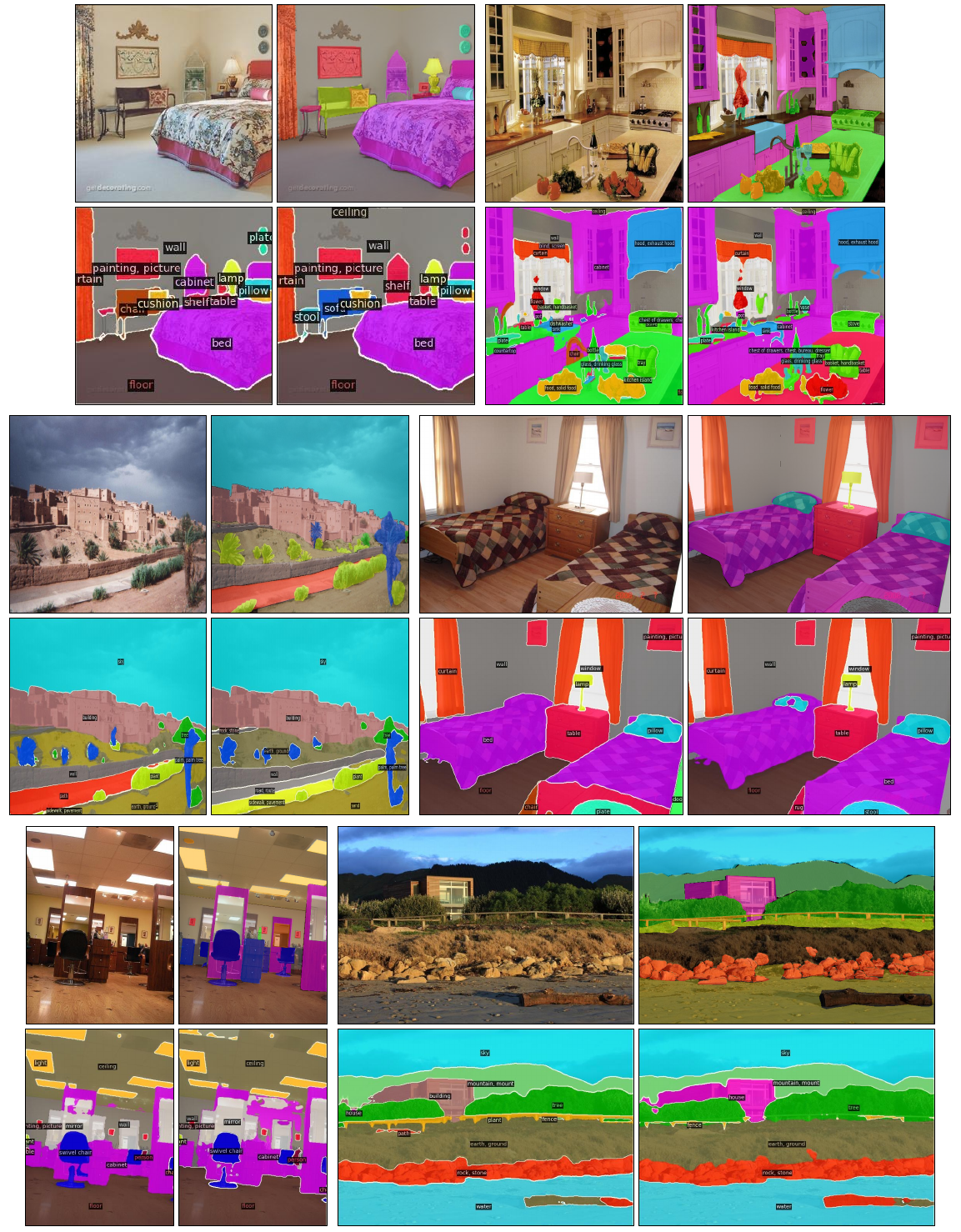}

    \vspace{-10pt}
    \caption{\textbf{Random Initialize vs. Mask-JEPA in Semantic Segmentation.} We compare the visualizations of semantic segmentation on the ADE20K dataset between a randomly initialized pixel decoder and transformer decoder (bottom-left) and those achieved by Mask-JEPA (bottom-right), using the Swin-T backbone in the Mask2Former model. Reference is made to the original image (upper-left) and the ground truth (upper-right). The results show that Mask-JEPA surpasses the baseline.}
    \vspace{-10pt}
    \label{fig:supp_output_sem}
\end{figure*}
\begin{figure*}[tp]
    \centering
    \includegraphics[width=0.95\linewidth]{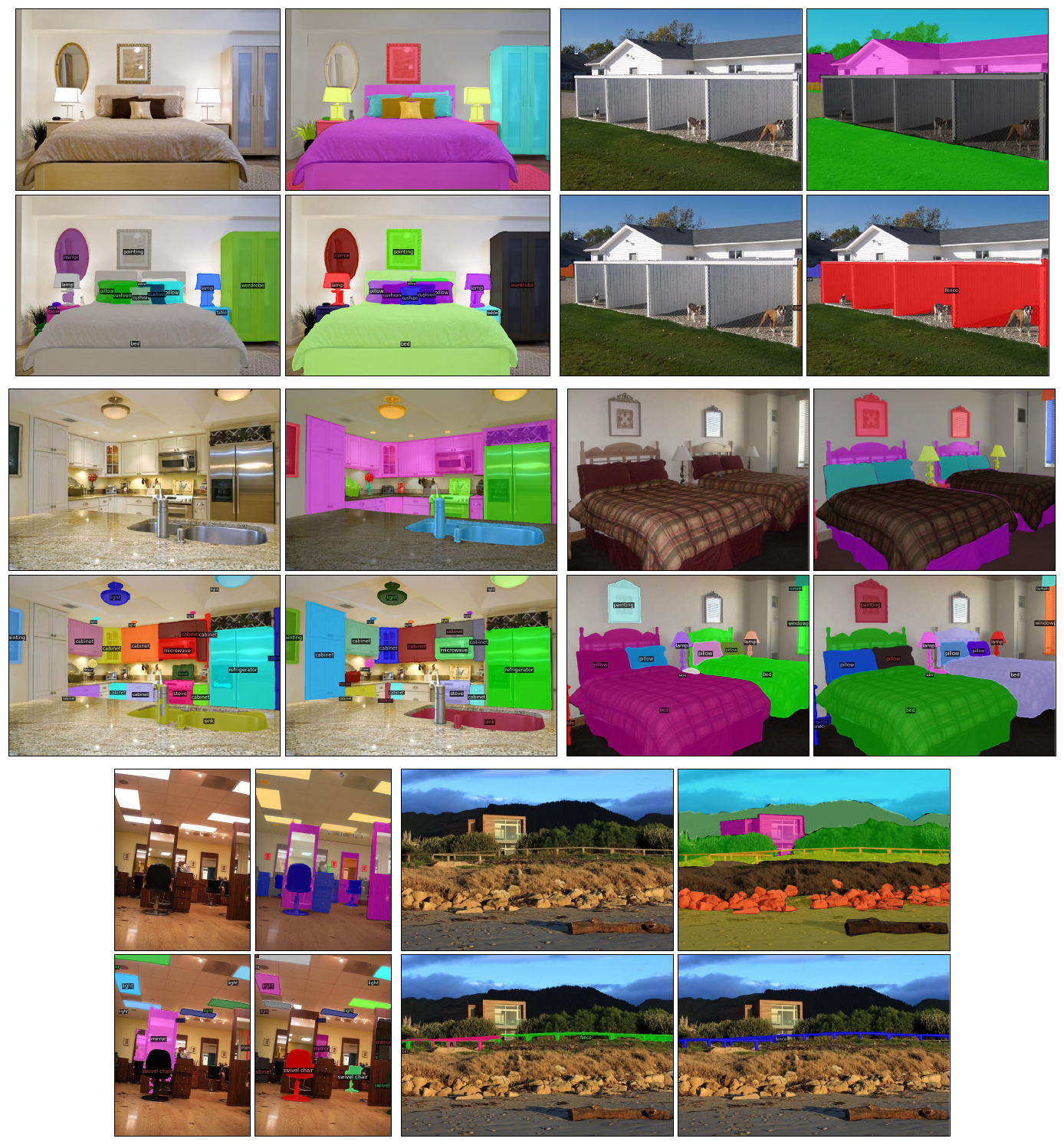}

    \vspace{-10pt}
    \caption{\textbf{Random Initialize vs. Mask-JEPA in Instance Segmentation.} We compare the visualizations of instance segmentation on the ADE20K dataset between a randomly initialized pixel decoder and transformer decoder (bottom-left) and those achieved by Mask-JEPA (bottom-right), using the Swin-T backbone in the Mask2Former model. Reference is made to the original image (upper-left) and the ground truth (upper-right). The results show that Mask-JEPA surpasses the baseline.}
    \vspace{-10pt}
    \label{fig:supp_output_ins}
\end{figure*}
We visualize and compare sample predictions from two configurations of the Mask2Former model in Figure~\ref{fig:supp_output_sem} and Figure~\ref{fig:supp_output_ins}: the standard Mask2Former model and the Mask2Former model trained with Mask-JEPA weights, both using a Swin-T~\cite{liu2021swin} backbone. These comparisons are made on two tasks: ADE20K \texttt{val} semantic segmentation in Figure~\ref{fig:supp_output_sem} (47.7 mIoU vs. 50.5 mIoU) and ADE20K \texttt{val} instance segmentation in Figure~\ref{fig:supp_output_ins} (27.2 AP vs. 28.5 AP).

\end{document}